\def\arxivversion{}
\documentclass{article}
\ifdefined\arxivversion\else
\pdftrailerid{}
\fi
\usepackage{iclr2027_conference,times}
\ifdefined\arxivversion\iclrfinalcopy\fi
\usepackage[hyphens]{url}
\usepackage{graphicx}
\usepackage{natbib}
\usepackage{caption}
\usepackage{ragged2e}
\usepackage{placeins}
\usepackage{needspace}
\usepackage{wrapfig}
\usepackage{algorithm}
\usepackage{algorithmic}
\usepackage{booktabs}
\usepackage{multirow}
\usepackage{amsmath}
\usepackage{mathtools}
\usepackage{amssymb}
\newtheorem{proposition}{Proposition}
\usepackage{xcolor}
\newsavebox{\worldscoretitlebox}

\DeclareRobustCommand{\worldscoretitle}{%
  \begingroup
  \sbox{\worldscoretitlebox}{%
    \normalfont\bfseries\fontsize{22}{24}\selectfont World4Scorer:}%
  \raisebox{-\dp\worldscoretitlebox}{%
    \includegraphics[width=\wd\worldscoretitlebox,
      height=\dimexpr\ht\worldscoretitlebox+\dp\worldscoretitlebox\relax]%
      {figures/title/world4score.pdf}%
  }%
  \endgroup
}

\usepackage{colortbl}
\usepackage{pifont}
\usepackage{xfp}
\definecolor{DeltaGreen}{HTML}{1B7837}
\definecolor{DeltaRed}{HTML}{B6321C}
\newlength{\deltaslot}
\newcommand{\panelgap}{14pt}
\newcommand{\tabdelta}[2]{%
  \makebox[\deltaslot][l]{${}_{\ifnum\fpeval{sign(#1-#2)}<0
    \color{DeltaRed}\fpeval{round(#1-#2,2)}%
  \else
    \color{DeltaGreen}+\fpeval{round(#1-#2,2)}%
  \fi}$}%
}
\newcommand{\prosedelta}[2]{\fpeval{round(abs(#1-#2),2)}}
\newcommand{\cmark}{{\footnotesize\textcolor{DeltaGreen}{\ding{51}}}}
\newcommand{\xmark}{{\footnotesize\textcolor{DeltaRed}{\ding{55}}}}
\definecolor{TableHeader}{HTML}{FFFFFF}
\definecolor{TableReference}{HTML}{F5F5F5}
\definecolor{TableDirect}{HTML}{E6F6FF}
\definecolor{TableVLA}{HTML}{F9F6C4}
\definecolor{TableWorld}{HTML}{FFEFF6}
\definecolor{TableBaseline}{HTML}{F5F5F5}
\definecolor{TableOurs}{HTML}{E9E9E9}
\definecolor{TableZebra}{HTML}{FAFAFC}
\definecolor{SubTone}{HTML}{454545}
\definecolor{ZebraA}{HTML}{FFFFFF}
\definecolor{ZebraB}{HTML}{E4E4EB}
\definecolor{RQpurple}{HTML}{9788C7}
\definecolor{RQblue}{HTML}{81A6CF}
\definecolor{RQgreen}{HTML}{6AB5A7}
\definecolor{RQamber}{HTML}{D0A16B}
\newcommand{\rqbadge}[2]{{\color{#1}#2}\,}

\newcommand{\epdmsSupportExpWrongHistory}{89.5} %
\newcommand{\epdmsSupportExpStaleHistory}{91.5} %
\newcommand{\historyCorrectWrongGap}{2.96} %
\newcommand{\historyCorrectWrongCILow}{2.63} %
\newcommand{\historyCorrectWrongCIHigh}{3.34} %
\newcommand{\historyCorrectStaleGap}{1.01} %
\newcommand{\historyCorrectStaleCILow}{0.86} %
\newcommand{\historyCorrectStaleCIHigh}{1.16} %
\newcommand{\navtestSize}{12{,}146}        %
\newcommand{\epdmsCopyPrev}{82.0}     %
\newcommand{\stateGtBefore}{1.87} %
\newcommand{\stateGtAfter}{1.51} %
\newcommand{\latentBankSize}{103{,}288}   %
\newcommand{\latentPairedCosMean}{0.94}   %
\newcommand{\latentPairedCosPFive}{0.80}  %
\newcommand{\latentDerangedCosMean}{0.649} %
\newcommand{\latentFloorCosMean}{0.649}   %
\newcommand{\latentAUC}{0.981}            %
\newcommand{\latentCohenD}{2.84}          %
\newcommand{\latentSameLogCos}{0.76}      %
\newcommand{\latentEffRank}{49}           %
\newcommand{\ecFlipPairs}{1939}      %
\newcommand{\ecRegressPairs}{18}     %
\newcommand{\ecFlipHeadingDrop}{32\%} %
\newcommand{\ecFlipLateralDrop}{21\%} %
\newcommand{\ecFlipPathDrop}{11\%}   %
\newcommand{\figTwoNumLogs}{136}       %
\newcommand{\figTwoNumCandidates}{64}  %
\newcommand{\epdmsSupportExp}{89.9}       %
\newcommand{\epdmsSupportExpEC}{92.5}     %

\newcommand{\ecLamQuarter}{92.42}  %
\newcommand{\ecLamHalf}{92.47}     %
\newcommand{\ecLamOne}{92.49}      %
\newcommand{\ecLamTwo}{92.45}      %
\newcommand{\ecLamBand}{0.07}      %

\newcommand{\epdmsReleasedSameHost}{90.66}
\newcommand{\epdmsReleasedState}{92.27}
\newcommand{\stateGainReleased}{+1.61}
\newcommand{\epdmsReleasedSameHostOne}{90.7}

\newcommand{\sharedStateEffRank}{17.9}
\newcommand{\sharedStateWithinShare}{75.8}

\newcommand{\sharedStateKnnAgree}{80.2}
\newcommand{\sharedStateKnnTraj}{76.8}
\newcommand{\sharedStateKnnShuffled}{52.0}
\newcommand{\sharedStateAboveTraj}{98}
\newcommand{\sharedStateShownPcaLow}{64}   %
\newcommand{\sharedStateShownPcaHigh}{75}
\newcommand{\sharedStateShownMinGroup}{15} %

\newcommand{\ecCeilingReleased}{92.66}    %

\newcommand{\humanEcPass}{90.1}       %
\newcommand{\humanEcInPool}{16.5}     %
\newcommand{\pdmsUnifiedFullBestMain}{94.0} %
\newcommand{\vTwoRawUnifiedNavTwoOne}{91.4}
\newcommand{\vTwoStateUnifiedNavTwoOne}{93.0}

\newcommand{\ecUnifiedNavTwoOffOne}{77.0}
\newcommand{\ecUnifiedNavTwoOnOne}{92.8}
\newcommand{\epUnifiedNavTwoOffOne}{90.0}
\newcommand{\epUnifiedNavTwoOnOne}{89.8}

\newcommand{\gmacsOurs}{349.9}          %
\newcommand{\gmacsOursHead}{0.74}       %
\newcommand{\gmacsOursTrue}{927.1}       %
\newcommand{\gmacsOursAttention}{577.2}  %
\newcommand{\gmacsOursEnc}{349.1}        %
\newcommand{\backboneShareHead}{0.08\%} %
\newcommand{\visualTokens}{15{,}744}     %
\newcommand{\latencyOursMs}{106.5}       %
\newcommand{\latencyOursHeadMs}{1.4}     %
\newcommand{\latencyOursTfMs}{75.6}      %
\newcommand{\latencyOursHalfMs}{23.0}    %
\newcommand{\latencyOursHalfEncMs}{20.2} %
\newcommand{\latencyStateTermMs}{125.4}  %
\newcommand{\paramsSubmittedM}{35.52}      %
\newcommand{\trunkSharedParams}{4{,}504{,}320} %
\newcommand{\trunkQKUnreached}{524{,}288}      %
\newcommand{\trunkQKUnreachedShare}{11.6\%}    %
\newcommand{\setDrawPerStep}{16}           %
\newcommand{\setValidMean}{54.5}           %
\newcommand{\setSeenShare}{30.8\%}         %
\newcommand{\setAbsentScenes}{310}         %
\newcommand{\setQuotaShortShare}{10.7\%}   %

\newcommand{\rSixMatchedRegret}{5.14}
\newcommand{\rSixMatchedRegretCI}{[4.66,\,5.62]}
\newcommand{\rSixSceneMeanRegret}{28.23}
\newcommand{\rSixSceneMeanRegretCI}{[25.97,\,30.50]}
\newcommand{\rSixSceneMeanDelta}{+23.09}
\newcommand{\rSixSceneMeanCI}{[+20.98,\,+25.17]}
\newcommand{\rSixDerangedRegret}{19.86}
\newcommand{\rSixDerangedRegretCI}{[18.65,\,21.08]}
\newcommand{\rSixDerangedDelta}{+14.72}
\newcommand{\rSixDerangedCI}{[+13.72,\,+15.75]}
\newcommand{\rSixMatchedSelected}{94.19}

\newcommand{\rSixMeanSelected}{71.10}

\newcommand{\rSixRepeatSelected}{79.73}
\newcommand{\rSixRepeatSelectedCI}{[78.64,\,80.85]}
\newcommand{\rSixRepeatMin}{77.44}
\newcommand{\rSixRepeatMax}{81.99}
\newcommand{\rSixRepeatDrop}{14.47}
\newcommand{\rSixRepeatDropCI}{[13.65,\,15.29]}

\newcommand{\ecFailCount}{2{,}717}    %
\newcommand{\ecPairCount}{10{,}040}   %
\newcommand{\ecTradeShare}{71.3\%}    %
\newcommand{\ecTieShare}{0.7\%}       %
\newcommand{\ecPoolShare}{27.9\%}     %
\newcommand{\ecTradeRepair}{97.6\%}   %
\newcommand{\ecTieRepair}{95\%}
\newcommand{\ecPoolRepair}{3.8\%}
\newcommand{\ecPoolQFour}{88.7\%}     %
\newcommand{\ecResidualPool}{91.7\%}  %

\newcommand{\rSixDerangedFlipRate}{100\%}         %
\newcommand{\rSixDerangedOrderAcc}{49.9\%}       %

\newcommand{\aThirtyRate}{2.6}        %
\newcommand{\aThirtyVram}{14}         %
\newcommand{\aThirtyFullHours}{1.3}   %

\newcommand{\vOneVTwoRho}{0.69}          %
\newcommand{\vOneVTwoRhoNoEc}{0.96}      %
\newcommand{\vOneVTwoArgmaxDisagree}{43\%}     %
\newcommand{\vOneVTwoArgmaxDisagreeNoEc}{3\%}  %
\newcommand{\vOneVTwoEcShare}{98\%}      %
\newcommand{\vOneVTwoVOneCost}{1.3}      %
\newcommand{\vOneVTwoEcFailShare}{99.7\%}%
\newcommand{\jointInfeasible}{38\%}      %
\newcommand{\jointInfeasibleQFour}{57\%} %
\newcommand{\epEcCorr}{$+0.07$}          %
\newcommand{\vOneVTwoTieMismatch}{39}                 %
\newcommand{\vOneVTwoEcRawDelta}{50.4}                %
\newcommand{\vOneVTwoEpRawDelta}{2.7}                 %
\newcommand{\vOneVTwoEpSaturated}{43\%}               %
\newcommand{\jointInfeasibleMedian}{30\%}             %
\newcommand{\jointInfeasibleWithinScene}{23\%}        %

\newcommand{\bankDim}{384}                 %

\newcommand{\selWeightsVOne}{(1,1,0,5,5,2)}              %
\newcommand{\selWeightsNavTwo}{(10,13,6,14,15,2.1)}      %

\newcommand{\pdmsLeanjcurLast}{92.70} %
\newcommand{\pdmsLeanjzeroLast}{93.01} %
\newcommand{\pdmsLeanjfullLast}{93.96} %

\newcommand{\permPdmControl}{0.728}
\newcommand{\permPdmClover}{0.647}
\newcommand{\permPdmAll}{0.461}
\newcommand{\permRandomPick}{0.524}
\newcommand{\permEvalScenes}{511}
\newcommand{\permOrderControl}{64.2\%}
\newcommand{\permOrderClover}{55.6\%}
\newcommand{\permOrderAll}{51.5\%}

\newcommand{\xlinDerangedDelta}{+14.11}
\newcommand{\xlinDerangedCI}{[+13.01,\,+15.19]}
\newcommand{\xlinOrderMatched}{72.5\%}
\newcommand{\xlinOrderDeranged}{50.2\%}
\newcommand{\xlinRegretMatched}{5.64}

\newcommand{\pdmSimPoses}{40}          %
\newcommand{\pdmSimInterval}{0.1}      %
\newcommand{\pdmSimHorizon}{4}         %
\newcommand{\pdmObsSteps}{51}          %

\newcommand{\bTwoDRouteTargetDist}{20}  %
\newcommand{\bTwoDRouteTargetScanFrames}{15{,}589}  %
\newcommand{\bTwoDOurs}{73.27}          %
\newcommand{\bTwoDSR}{46.36}            %
\newcommand{\bTwoDFutWeight}{0.1}       %
\newcommand{\bTwoDFutGradScale}{0.05}   %
\newcommand{\bTwoDEpochs}{15}          %
\newcommand{\bTwoDBatch}{24}           %
\newcommand{\bTwoDLR}{$7.5\times10^{-5}$} %

\newcommand{\navtestTokens}{12{,}146}   %

\newcommand{\cubeCandidates}{300}        %
\newcommand{\cubeElite}{30}              %
\newcommand{\cubeHorizonBlocks}{5}       %
\newcommand{\cubeFrameskip}{5}           %
\newcommand{\cubeBudget}{50}             %
\newcommand{\cubeGoalOffset}{25}         %
\newcommand{\cubeToleranceCm}{4}         %
\newcommand{\cubeEpisodes}{50}           %
\newcommand{\cubeSeedMain}{42}           %
\newcommand{\cubeItersAppD}{10}          %
\newcommand{\cubeBlendC}{0.3}            %
\newcommand{\cubeTuneSeeds}{200--203}    %
\newcommand{\cubeReportSeeds}{11}        %
\newcommand{\cubeAnchorTenSeedMain}{72}  %
\newcommand{\cubeOursTenSeedMain}{80}    %
\newcommand{\cubePLDM}{65}     %
\newcommand{\cubeGCIQL}{64}    %
\newcommand{\cubeGCIVL}{56}    %

\newcommand{\cubeAnchorTenMean}{68.91}   %
\newcommand{\cubeOursTenMean}{73.64}     %
\newcommand{\cubeDeltaTen}{+4.73}        %
\newcommand{\cubeDeltaTenCI}{[+2.18,\,+7.27]} %
\newcommand{\cubeSeedsUpTen}{9}          %
\newcommand{\cubeSeedsTieTen}{1}         %
\newcommand{\cubeSeedsDownTen}{1}        %
\newcommand{\cubeTenEpisodes}{550}
\newcommand{\cubeAnchorTenSucc}{379}
\newcommand{\cubeOursTenSucc}{405}
\newcommand{\cubeTenRescued}{39}
\newcommand{\cubeTenLost}{13}

\newcommand{\figfourNullGainDrivable}{+0.005}  %
\newcommand{\figfourNullGainFar}{+0.041}       %

\newcommand{\queryRoadConsistency}{0.750}
\newcommand{\queryDynamicConsistency}{0.639}
\newcommand{\queryStaticConsistency}{0.606}

\newcommand{\vOneVTwoDdcShare}{3\%}          %
\newcommand{\vOneVTwoRestShare}{$-1\%$}      %
\newcommand{\vOneVTwoEpShare}{$-0.7\%$}      %
\newcommand{\ecNoPairFrames}{2{,}106}        %
\newcommand{\decodeTrainPairs}{6{,}000}      %
\newcommand{\decodeGridCosGTwo}{0.681}       %
\newcommand{\decodeGridCosGOne}{0.707}       %
\newcommand{\decodeGridCosNull}{0.701}       %
\newcommand{\paramsViTLM}{321}               %
\newcommand{\paramsViTHM}{888}               %
\newcommand{\latencyHalfSpeedup}{4.6}        %
\newcommand{\trainHours}{40}                 %
\newcommand{\bTwoDRouteScanClips}{60}        %
\newcommand{\bTwoDTurnSignLeft}{97.4\%}      %
\newcommand{\bTwoDTurnSignRight}{100.0\%}    %
\newcommand{\bTwoDRouteEndShare}{10.5\%}     %

\newcommand{\mrgDAWN}{97.3 & 92.0 & 99.7 & 99.1 & 84.3 & 87.4 & 96.0 & 96.6 & 100.0 & 96.0 & 98.3 & 85.5 & 89.1 & 83.2}

\newcommand{\mrgDiffusionDrive}{98.2 & 96.2 & 99.8 & 99.5 & 82.2 & 87.4 & 94.7 & 97.3 & 100.0 & 97.0 & 98.3 & 87.4 & 88.1 & 88.2}
\newcommand{\mrgDiffusionDriveVTwo}{97.7 & 96.6 & 99.8 & 99.2 & 87.5 & 88.9 & 94.8 & 97.2 & 99.9 & 96.0 & 97.8 & 91.0 & 91.2 & 87.5}
\newcommand{\mrgDiscreteWAM}{98.5 & 98.2 & 99.8 & 99.7 & 88.7 & 90.5 & 95.3 & 97.9 & 100.0 & 97.2 & 98.3 & 78.1 & 92.2 & 90.4}

\newcommand{\mrgDriveFuture}{98.8 & 99.1 & 99.9 & 99.6 & 84.2 & 86.6 & 95.4 & 98.4 & 100.0 & 96.4 & 98.3 & 74.8 & 90.7 & 89.9}
\newcommand{\mrgDriveJEPA}{98.4 & 98.6 & 99.8 & 99.7 & 91.4 & 91.4 & 96.2 & 97.8 & 99.9 & 97.6 & 97.9 & 76.9 & \underline{93.7} & 90.8}

\newcommand{\mrgDriveVLAWZero}{98.5 & 99.1 & 99.7 & 98.0 & 83.3 & 86.4 & 95.3 & 98.1 & 99.3 & 93.2 & 97.9 & 58.9 & 90.2 & 86.1}

\newcommand{\mrgDriveWorldVLA}{98.6 & 99.1 & 99.8 & 99.6 & 85.9 & 87.4 & 96.1 & 97.9 & 100.0 & 97.0 & 97.8 & 78.6 & 91.3 & 86.8}
\newcommand{\mrgDrivorReleased}{99.3 & 99.2 & 99.8 & 99.7 & 89.9 & 87.4 & 96.7 & 98.8 & 100.0 & 95.8 & 98.0 & 75.5 & \underline{93.7} & 90.7}

\newcommand{\mrgExploreVLA}{98.8 & 96.2 & 99.8 & 99.6 & 83.5 & 87.1 & 96.5 & 98.2 & 99.9 & 97.8 & 98.3 & 86.8 & 90.4 & 88.8}

\newcommand{\mrgGraphWorld}{98.4 & 98.8 & 99.1 & 99.1 & 83.2 & 85.9 & 95.5 & 97.9 & 100.0 & 96.0 & 97.8 & 74.6 & 90.1 & 89.5}

\newcommand{\mrgLWDrive}{98.8 & 98.4 & 99.7 & 99.0 & 87.3 & 90.3 & 96.2 & 98.6 & 99.8 & 96.3 & 97.9 & 73.3 & 92.0 & 89.6}

\newcommand{\mrgOursOff}{99.3 & 99.2 & 99.6 & 99.7 & 91.3 & 90.0 & 96.0 & 98.8 & 100.0 & 96.1 & 97.8 & 77.0 & \textbf{94.0} & \textcolor{black}{\underline{91.4}}}
\newcommand{\mrgOursOn}{99.3 & 99.2 & 99.6 & 99.7 & -- & 89.8 & -- & 98.7 & -- & 96.2 & 97.8 & 92.8 & -- & \textcolor{black}{\textbf{93.0}}}

\newcommand{\mrgReCogDrive}{98.3 & 97.7 & 99.8 & 98.4 & 87.3 & 89.9 & 94.9 & 97.4 & 100.0 & 90.9 & 97.8 & 27.6 & 90.8 & 82.7}
\newcommand{\mrgSafeDrive}{99.5 & 99.0 & 99.9 & 99.7 & 84.3 & 88.6 & 97.2 & 98.9 & 100.0 & 97.5 & 98.2 & 81.9 & 91.6 & 87.5}

\newcommand{\mrgTransFuser}{97.8 & 92.1 & 99.8 & 99.2 & 79.2 & 87.5 & 92.8 & 96.6 & 100.0 & 95.7 & 98.3 & 88.7 & 84.0 & 83.9}

\newcommand{\mrgUniTeD}{99.1 & 97.2 & 99.9 & 99.6 & 84.1 & 86.9 & 96.6 & 98.5 & 100.0 & 98.4 & 99.9 & 87.3 & 90.2 & 90.1}
\newcommand{\mrgWAMFlow}{99.5 & 98.4 & 99.9 & 99.4 & 82.3 & 84.6 & 97.0 & 99.3 & 99.7 & 96.9 & 96.9 & 33.8 & 90.3 & 84.9}
\newcommand{\mrgWoTE}{98.5 & 96.8 & 99.8 & 98.8 & 81.9 & 86.1 & 94.9 & 97.9 & 99.9 & 95.5 & 98.3 & 82.9 & 88.3 & 87.6}

\newcommand{\mrgWorldRFT}{97.8 & 96.5 & 99.8 & 99.5 & 81.7 & 88.5 & 94.0 & 97.0 & 100.0 & 97.4 & 98.1 & 69.1 & 87.8 & 86.7}

\title{\worldscoretitle\\
{\Large Outcome-Grounded World Modeling\\for Autonomous Driving}}
\ifdefined\arxivversion
\usepackage{fontawesome5}
\usepackage{twemojis}
\definecolor{LinkAccent}{HTML}{6B4FD8}
\newcommand{\linkPage}{https://guobapei.github.io/World4Scorer/}
\newcommand{\linkHF}{https://huggingface.co/pei2333/World4Scorer}
\newcommand{\linkCode}{https://github.com/GuobaPei/World4Scorer}
\newcommand{\hithink}{HiThink Research}
\makeatletter
\def\@maketitle{\vbox{\hsize\textwidth\centering
{\LARGE\sc \@title\par}\vskip 9pt
\lhead{Preprint}%
{\parskip=0pt\@author\par}%
\vskip 0.05in}}
\makeatother
\AtBeginDocument{\hypersetup{pdftitle={World4Scorer: Outcome-Grounded World Modeling for Autonomous Driving}, pdfauthor={Jieyuan Pei, Meiyi Lu, Sining Ang, Yubo Zhao, Zhangyi Hu, Mingwei Xu, Haokai Ding, Wei Li, Zihan You, Jianwei Zheng, Li Yu, Yifeng Pan, Ji Tao, Rongjunchen Zhang, Yan Wang}}}
\DeclareRobustCommand{\worldscoretitle}{%
  \begingroup
  \sbox{\worldscoretitlebox}{\normalfont\bfseries\fontsize{22}{24}\selectfont World4Scorer}%
  \raisebox{-\dp\worldscoretitlebox}{\includegraphics[width=\wd\worldscoretitlebox,
      height=\dimexpr\ht\worldscoretitlebox+\dp\worldscoretitlebox\relax]{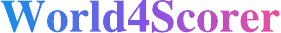}}%
  \endgroup}
\title{\worldscoretitle\\{\large Outcome-Grounded World Modeling for Autonomous Driving}}
\newcommand{\authsep}{\nobreak\hspace{0.3em plus 0.1em minus 0.12em}\nobreak}
\newcommand{\arxivauthors}{%
{\fontsize{8.6}{10.4}\selectfont\bfseries Jieyuan~Pei$^{1,2,3*\ddagger}$\hspace{0.3em plus 0.1em minus 0.12em} Meiyi~Lu$^{4*}$\hspace{0.3em plus 0.1em minus 0.12em} Sining~Ang$^{5}$\hspace{0.3em plus 0.1em minus 0.12em} Yubo~Zhao$^{6}$\hspace{0.3em plus 0.1em minus 0.12em} Zhangyi~Hu$^{3}$\hspace{0.3em plus 0.1em minus 0.12em}
Mingwei~Xu$^{7}$\hspace{0.3em plus 0.1em minus 0.12em} Haokai~Ding$^{8}$\\ Wei~Li$^{9}$\authsep Zihan~You$^{1,10}$\authsep Jianwei~Zheng$^{9}$\authsep Li~Yu$^{11}$\authsep Yifeng~Pan$^{11}$\authsep Ji~Tao$^{11}$\authsep Rongjunchen~Zhang$^{2}$\authsep Yan~Wang$^{1\dagger}$\par}\vskip 3pt
{\fontsize{7.6}{9}\selectfont 
\mbox{$^{1}$Institute for AI Industry Research (AIR), Tsinghua University}\hspace{0.7em}\mbox{$^{2}$\hithink}\\
\mbox{$^{3}$The Hong Kong University of Science and Technology (Guangzhou)}\hspace{0.7em}\mbox{$^{4}$Zhejiang University}\\
\mbox{$^{5}$University of Science and Technology of China}\hspace{0.7em}\mbox{$^{6}$SMBU}\hspace{0.7em}\mbox{$^{7}$University of Washington}\\
\mbox{$^{8}$Mohamed bin Zayed University of Artificial Intelligence}\hspace{0.7em}\mbox{$^{9}$Zhejiang University of Technology}\hspace{0.7em}\mbox{$^{10}$Southeast University}\\
\mbox{$^{11}$Changan Automobile}\hspace{0.7em}\hspace{0.9em}\mbox{$^{*}$Equal contribution}\hspace{0.7em}\mbox{$^{\dagger}$Corresponding author}\hspace{0.7em}\mbox{$^{\ddagger}$Work done during an internship}\par}\vskip 1pt
{\centering\fontsize{8.4}{8.8}\selectfont\bfseries\color{LinkAccent}\hypersetup{pdfborder={0 0 0}} \href{\linkPage}{\raisebox{-0.1ex}{\faHome}\,Project Page}\hspace{3.2em}\href{\linkHF}{\raisebox{-0.12ex}{\twemoji[height=0.95em]{1f917}}\,Hugging Face}\hspace{3.2em}\href{\linkCode}{\raisebox{-0.1ex}{\faGithub}\,Code}\par}}
\author{\arxivauthors}

\else
\author{Anonymous authors\\
Paper under double-blind review}
\fi

\usepackage{hyperref}   %

\begin{document}
\maketitle
\ifdefined\arxivversion\lhead{Preprint}\fi

\begin{abstract}
Autonomous driving requires choosing a safe and efficient plan as surrounding
traffic evolves. Generate-and-select planners propose multiple trajectories and
score them for execution, and they have outperformed representative
direct-prediction baselines on NAVSIM. Their scorer must compare plans that
were never executed. Driving logs record the future of only the executed
trajectory, so matching the logged future can leave predictions for the
alternatives unconstrained; a simulator, in contrast, can label the outcome of
every candidate. We introduce World4Scorer, which builds the scorer as a
trajectory-conditioned JEPA-style predictor: it predicts a state for each
candidate and reads the candidate's scores from that state. Simulator outcome
labels supervise the states of all candidates, and the observed future of the
executed trajectory anchors the predictor to real scene evolution. Because one
predictor produces every candidate's state, the anchor can constrain shared parameters used to score unexecuted plans, while the future itself is needed only
during training. Generated candidates mostly score well, so a scene-matched
bank adds low-scoring plans to the outcome supervision; framewise choices can conflict, so inertial re-ranking keeps consecutive selections consistent.
World4Scorer achieves state-of-the-art NAVSIM-v2 performance and a strong
adapted-system result on closed-loop Bench2Drive. With the LeWM world model and
planning budget fixed, outcome-based scoring also improves manipulation
planning on the OGBench-Cube benchmark.
\end{abstract}

\suppressfloats[t]
\begin{figure}[t]
\centering
\includegraphics[width=0.88\textwidth]{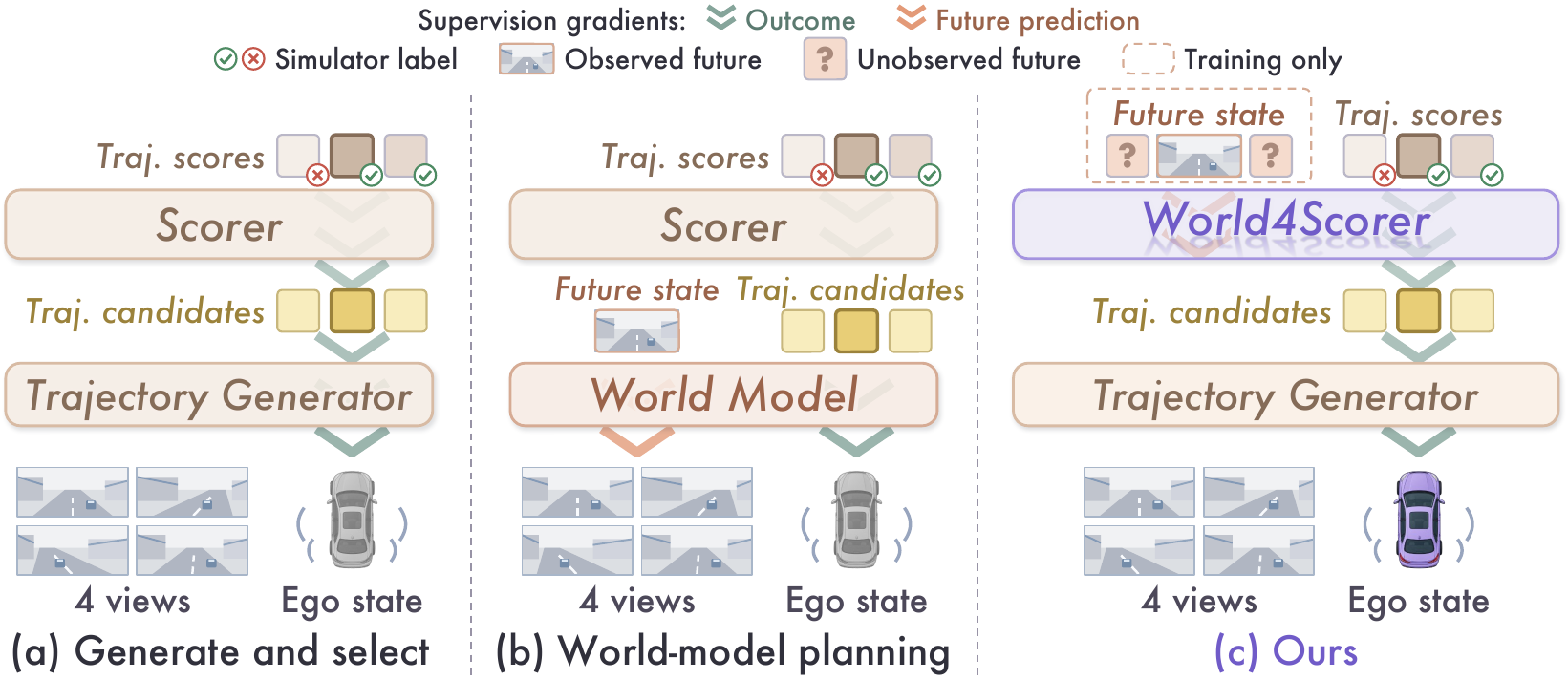}
\captionsetup{font=small,skip=3pt,justification=RaggedRight,singlelinecheck=false}
\caption{\textbf{Gradient routing in generator--scorer planning.}
(a) Outcome labels train the scorer.
(b) A world model adds future prediction, but only the executed trajectory has
an observed future.
(c) In World4Scorer, outcome labels supervise the state of every candidate,
and the observed future anchors the predictor.}
\label{fig:teaser}
\end{figure}

\section{Introduction}
\label{sec:intro}

Autonomous driving requires turning sensor observations into safe,
efficient, and comfortable motion. Direct trajectory-prediction methods
such as TransFuser~\citep{transfuser} and UniAD~\citep{uniad} map scene
features to a driving plan.
Recent generate-and-select planners have achieved stronger performance than
representative direct-prediction baselines on
NAVSIM~\citep{diffusiondrive,gtrs,drivor}. They use a trajectory generator
to propose alternatives and a learned scorer to choose one for execution
(Figure~\ref{fig:teaser}a). To benefit from these alternatives,
the scorer must identify which plan best balances safety, progress, and comfort.

Scoring a plan requires assessing its consequences as surrounding traffic
evolves. World models provide predictive features for planning~\citep{law,drivefuture}
and candidate evaluation~\citep{wote} (Figure~\ref{fig:teaser}b).
Their training faces a supervision gap: each driving log records the future
of the executed trajectory, while alternative trajectories have no corresponding
future observations. Matching the logged future can leave predictions for
unobserved alternatives unconstrained, even though these predictions decide which plan is selected.

Candidate outcomes fill this gap. Simulation can label whether each
alternative plan collides, leaves the drivable area, or makes progress,
without an observed visual future for that plan. When a predictor produces
the state from which each candidate is scored, these labels supervise the
state of every candidate, including plans that were never executed.
The observed future serves as an anchor: it supervises the state of the
executed trajectory and ties the shared predictor to real scene evolution.

We introduce World4Scorer, an \emph{outcome-grounded world modeling}
framework in which a trajectory-conditioned predictor is the scorer
(Figure~\ref{fig:teaser}c). Following the joint-embedding predictive
architecture (JEPA)~\citep{jepa,vjepa}, the predictor maps each candidate to
a state; score heads read candidate outcomes from this state, and a visual
readout predicts the future features of the executed trajectory. The core is
this shared predictor, trained by both candidate outcomes and the observed
future. As support, a perturbed candidate bank widens outcome coverage, and
training-free inertial re-ranking checks continuity with the previous plan.
\begin{itemize}\setlength{\itemsep}{1pt}\setlength{\parskip}{0pt}
\item \textbf{Outcome-grounded world modeling.} We build the candidate scorer
as a trajectory-conditioned predictor. Simulator outcomes supervise the
predicted state of every generated and bank candidate, including plans that
were never executed, while the observed future of the executed plan anchors
the shared predictor to real scene evolution.
\item \textbf{Complementary supervision.} Outcome labels reach every candidate;
the observed future reaches one trajectory yet constrains the shared predictor.
In controlled tests, an observed-future target improves planning but
a current-frame target does not, mismatched outcome labels degrade ranking,
and reassigning candidate states lowers PDMS in every trial.
\item \textbf{Evidence across benchmarks and domains.} World4Scorer achieves
state-of-the-art performance on NAVSIM-v2 and leads the compared methods on
NAVSIM-v1. Cross-setting tests extend these results: an adapted system is
strong in closed-loop Bench2Drive, and outcome-based scoring improves
manipulation with a frozen world model.
\end{itemize}

\section{Related Work}
\label{sec:related}

\paragraph{End-to-end autonomous driving.}
TransFuser~\citep{transfuser} and UniAD~\citep{uniad} regress a trajectory
from scene features. Candidate-based planners compare alternatives
from vocabularies or generators.
VADv2~\citep{vadv2} predicts probabilities over a fixed
vocabulary. DiffusionDrive~\citep{diffusiondrive} refines
anchors and selects by learned confidence. GTRS~\citep{gtrs} scores
vocabulary and generated candidates, and
DriveSuprim~\citep{drivesuprim} filters candidates from coarse to fine.
CLOVER~\citep{clover} trains its generator on evaluator-filtered
pseudo-experts and its scorer on evaluator sub-scores. DrivoR~\citep{drivor} learns proposals and
simulator-supervised scores with separate decoders.
We follow its generator--scorer framework and make the scorer a predictor
trained by simulator labels, anchored by observed futures.

\paragraph{Predictive representations for planning.}
Future prediction supports both trajectory generation~\citep{law,drivelaw,drivefuture}
and candidate evaluation~\citep{wote,safedrive,worldfourdrive}. Drive-JEPA~\citep{drivejepa} uses video-JEPA pretraining to provide predictive
features for planning. We focus on the supervision available during candidate
comparison: observed futures cover executed trajectories, whereas simulator
outcomes cover every candidate; World4Scorer trains one predictor with both.

\raggedbottom
\begin{figure*}[t]
\centering
\includegraphics[width=\textwidth]{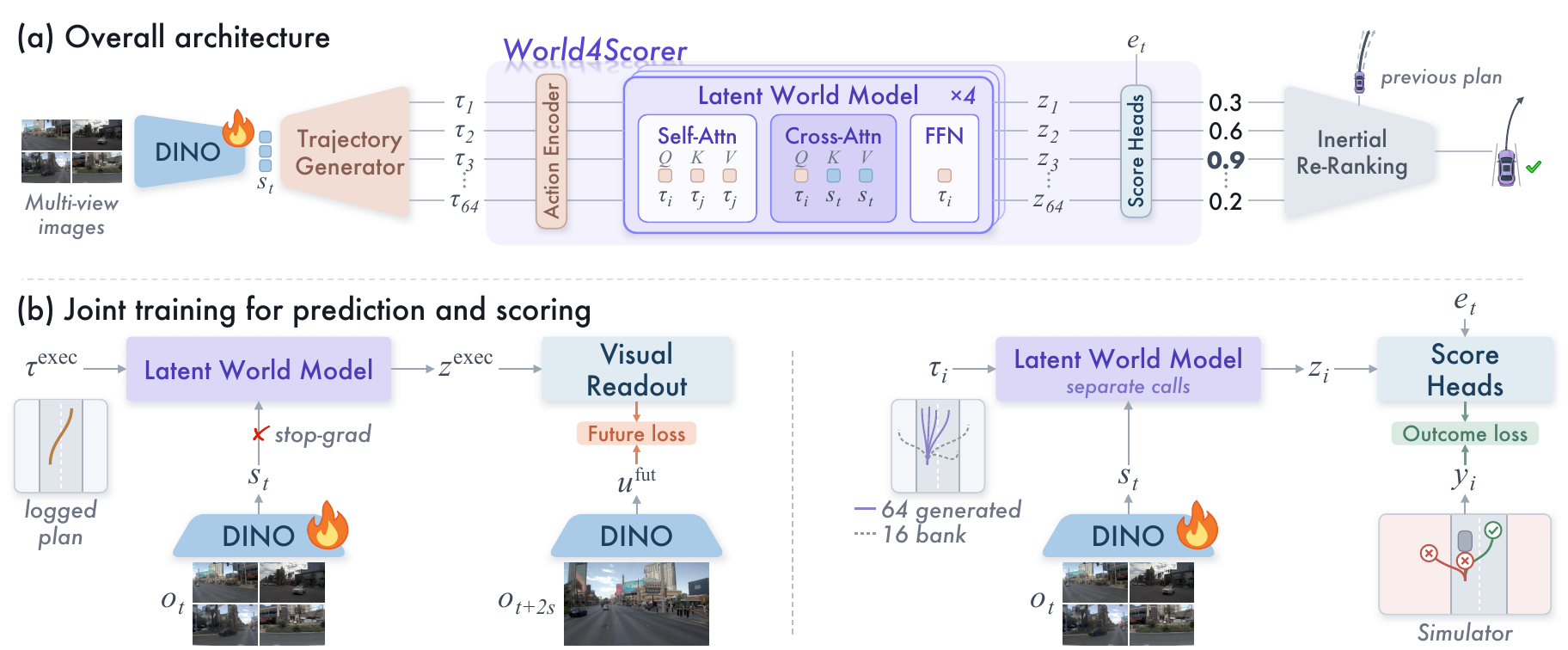}
\caption{\textbf{Overview of World4Scorer.}
(a) The trajectory generator proposes 64 candidates, scored by World4Scorer
before inertial re-ranking.
(b) The visual readout and score heads train the shared predictor: future observations supervise
visual prediction for the executed trajectory; outcome labels supervise
scoring for all generated candidates and 16 bank trajectories.}
\label{fig:model}
\end{figure*}

\Needspace{8\baselineskip}
\section{\textnormal{World4Scorer}}
\label{sec:method}

World4Scorer follows a generator--scorer framework: it proposes trajectories,
predicts a state for each candidate, and scores these states before inertial
re-ranking (Figure~\ref{fig:model}). We use ``world model'' in a decision-focused sense: latent states
carry candidate outcomes, not full futures.

\subsection{Overview}
\label{sec:method-overview}

At inference (Figure~\ref{fig:model}a), DINOv2~\citep{dinov2} with LoRA~\citep{lora}
encodes four camera views into scene tokens $s_t$. A transformer generates
$N=64$ trajectories, $\mathcal C_t=\{\tau_{t,i}\}_{i=1}^{N}$.
The predictor maps each trajectory query to a state using the scene tokens.
Score heads read each state together with the ego state. Inertial re-ranking
then adds compatibility with the previous plan.

At training time (Figure~\ref{fig:model}b), winner-take-all imitation supervises
the generator. Outcome labels supervise all 64 generated candidates and 16
trajectories sampled from a scene-matched bank. A separate query for the logged
trajectory predicts its observed visual future. These three query sets use the
same predictor, allowing visual prediction and outcome scoring to train shared parameters.

\subsection{Shared Prediction and Scoring}
\label{sec:method-base}

An action encoder $\phi$ maps each candidate trajectory to a query token.
The shared predictor $P$ (Latent World Model in Figure~\ref{fig:model})
processes all candidate queries together with the scene tokens:
\begin{equation}
q_{t,i}=\phi(\operatorname{sg}[\tau_{t,i}]),\quad
Q_t=[q_{t,1},\ldots,q_{t,N}],\quad
Z_t=P(Q_t,s_t).
\label{eq:shared-scoring}
\end{equation}
Its four transformer blocks exchange information among candidates through
self-attention and incorporate scene information through cross-attention,
followed by output normalization. Row $z_{t,i}$ of $Z_t$ is the predicted
state for candidate $i$, and $\operatorname{sg}$ stops gradients.

The score heads $H$ read each candidate state after adding the
encoded ego state $e_t$: $\boldsymbol\ell_{t,i}=H(z_{t,i}+e_t)$.
They predict collision avoidance (NC), drivable-area compliance (DAC),
driving direction (DDC), time-to-collision (TTC), progress (EP), and comfort (C).
Let $p_{t,i,m}=\sigma(\ell_{t,i,m})$,
$\mathcal G=\{\mathrm{NC,DAC,DDC}\}$, and
$\mathcal A=\{\mathrm{TTC,EP,C}\}$. The ranking score is
\begin{equation}
\hat J_{t,i}
=\sum_{m\in\mathcal G}\alpha_m\log p_{t,i,m}
+\log\!\left(\sum_{m\in\mathcal A}\beta_m p_{t,i,m}\right).
\label{eq:ranking-utility}
\end{equation}
Here $\alpha_m$ and $\beta_m$ are per-benchmark weights (Appendix~\ref{sec:appendix-training-spec}). The highest-scoring
candidate defines the framewise choice; inertial re-ranking further incorporates
previous-plan compatibility.

Detaching candidate coordinates prevents scoring gradients from reaching
the generator through its trajectory outputs. The scoring loss still trains
$\phi$, $P$, $H$, and the scene features.

\Needspace{8\baselineskip}
\subsection{Outcome-Grounded Predictive Learning}
\label{sec:method-predictive}

Only the executed trajectory has an observed future; generated and bank
candidates lack future observations but have outcome labels. Outcome labels
supervise every candidate state, and the observed future anchors the predictor
through the executed trajectory (Figure~\ref{fig:model}b).

\noindent\textbf{Trajectory-conditioned JEPA.}
Following JEPA~\citep{jepa,vjepa}, the predictor takes scene tokens and the
logged trajectory query. Its target is the frozen DINOv2 embedding of the
observed front view two seconds after the current frame, mean-pooled to
384 dimensions:
\begin{equation}
z_t^{\rm exec}=P([\phi(\tau_t^{\rm exec})],\operatorname{sg}[s_t]),
\qquad u_t^{\rm fut}=f_{\rm DINO}(o_{t+2\mathrm{s}}).
\label{eq:executed-state}
\end{equation}
The linear visual readout $g$ predicts this feature with a cosine loss:
\begin{equation}
\mathcal L_{\rm fut}=\frac1M\sum_{n=1}^{M}
\left[1-\operatorname{sim}_{\cos}
\bigl(g(z_n^{\rm exec}),u_n^{\rm fut}\bigr)\right].
\label{eq:factual-jepa}
\end{equation}
Here $n$ indexes $M$ scenes. With scene tokens detached, this loss updates
$\phi$, $P$, and $g$; at deployment, the predictor needs only current observations and candidate trajectories.
Because every query passes through the same $P$, this single target also
constrains the parameters that produce each candidate's state. The self-attention
query and key projections receive no gradient from this single-query loss and
are trained by the outcome losses alone (Appendix~\ref{sec:appendix-training-spec}).

\noindent\textbf{Learning from candidate outcomes.}
Outcome labels $y_i$ supervise $H(z_i+e)$:
\begin{equation}
\mathcal L_{\rm out}
=\mathcal L_{\rm score}+\lambda_{\rm bank}\mathcal L_{\rm bank}
=\sum_i w_i\,\ell_{\rm out}\bigl(H(z_i+e),y_i\bigr),
\label{eq:outcome-predictive-learning}
\end{equation}
where $i$ covers generated and bank candidates, $\ell_{\rm out}$ uses each set's
loss components, and $w_i$ includes averaging and bank weighting
(Section~\ref{sec:method-candidates}). Outcome supervision thus reaches every
candidate state, including those without visual targets, and requires
distinctions in safety, progress, and comfort. A scene supplies outcome labels for up to 80 candidates and a future target for one.

\noindent\textbf{Why future supervision still matters.}
Models that fit the available outcome labels may still disagree on other
candidates~\citep{valueequivalence}, including which one to
select~\citep{decisionfocusedranking}. We bound this variation with a local generalized Gauss--Newton model
around a fitted $\theta_0$, keeping each predictor call in its own context.

\begin{proposition}[Future supervision and decision stability]
\label{prop:anchor-contrast}
Let $\mu,\lambda_{\rm fut},\epsilon>0$, and let the executed-query readout be differentiable and nonzero at $\theta_0$. Let $G_{\rm out}$ (with a trust-region term $\mu I$) and $G_{\rm fut}$ be the generalized Gauss--Newton curvatures of the BCE outcome and cosine future losses at $\theta_0$, and let
$\rho_{ij}(G)$ be the largest first-order change $|c_{ij}^\top\delta|$ over
$\delta^\top G\delta\le\epsilon^2$. Then
\begin{equation}
\rho_{ij}^{\rm out+fut}=\rho_{ij}(G_{\rm out}+\lambda_{\rm fut}G_{\rm fut})
\le\rho_{ij}(G_{\rm out})=\rho_{ij}^{\rm out},
\label{eq:complementary-supervision}
\end{equation}
strictly if and only if $J_{\rm fut}G_{\rm out}^{-1}c_{ij}\ne0$, where
$c_{ij}=\nabla_\theta(\hat J_i-\hat J_j)$ and $J_{\rm fut}$ is the Jacobian of
the normalized readout. If $\hat J_i-\hat J_j>\rho_{ij}^{\rm out+fut}$, the linearized ordering $i\succ j$ holds throughout.
\end{proposition}

\noindent
\begin{minipage}[t]{0.44\columnwidth}
\vspace{0pt}
Figure~\ref{fig:identifiability} shows how future constraints can remove
local disagreement on candidate choice. A smaller variation range makes a
candidate's lead harder to reverse within the allowed neighborhood. The future
loss does not constrain every unexecuted candidate uniformly: it narrows a
comparison only along directions that shared parameters couple to the executed
query's readout. Appendix~\ref{sec:appendix-anchor-contrast} proves this local result;
Section~\ref{sec:rq-components} evaluates the learning effects predicted by this mechanism.
\end{minipage}\hfill
\begin{minipage}[t]{0.53\columnwidth}
\vspace{0pt}
\centering
\includegraphics[width=\linewidth]{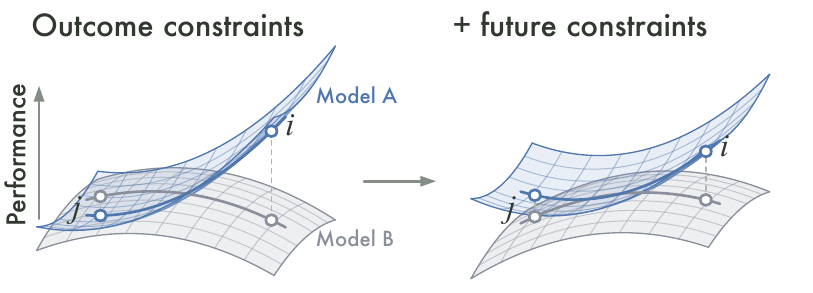}
\captionsetup{font=small,skip=3pt,justification=RaggedRight,singlelinecheck=false}
\captionof{figure}{\textbf{Complementary supervision.}
Two score surfaces fit the outcome labels but favor different candidates.
In this example, future constraints remove that disagreement while allowing
different predictions elsewhere.}
\label{fig:identifiability}
\end{minipage}
\par\smallskip

\Needspace{17\baselineskip}
\subsection{Outcome Supervision across Candidates}
\label{sec:method-candidates}

\noindent
\begin{minipage}[t]{0.60\columnwidth}
\vspace{0pt}
\setlength{\emergencystretch}{1em}
Imitation training yields mostly expert-like, high-scoring candidates
(light bars in Figure~\ref{fig:bank}), so adverse outcomes such as collisions
and drivable-area violations form a small share of the training examples.

Following CLOVER~\citep{clover}, we use a scene-matched bank of
perturbed trajectories with precomputed simulator labels. The bank covers a
broader score range, including unsafe plans (dark bars).
We sample $\setDrawPerStep{}$ bank trajectories per step to complement
the generated candidates.
\end{minipage}\hfill
\raisebox{30pt}[\height][\dimexpr\depth-30pt\relax]{%
\begin{minipage}[t]{0.36\columnwidth}
\vspace{0pt}
\centering
\includegraphics[width=\linewidth]{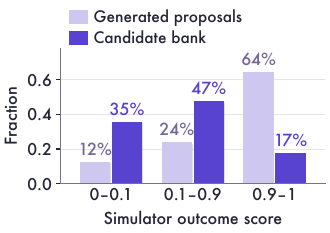}
\captionsetup{font=small,skip=3pt,justification=RaggedRight,singlelinecheck=false}
\captionof{figure}{\textbf{Candidate bank.}
The bank adds low-scoring plans.}
\label{fig:bank}
\end{minipage}%
}
\par\vspace{-0.42\baselineskip}\noindent
Both sets use separate calls to the same $\phi$, $P$, and $H$.
For generated candidates, $\mathcal L_{\rm score}$ sums binary cross-entropy
over six components, using soft targets for progress.
For bank candidates, $\mathcal L_{\rm bank}$ averages binary cross-entropy
over five non-progress components and adds $0.1$ times the progress L1 loss.
Each component loss is averaged over scenes and candidates in its set.

Winner-take-all L1 imitation, $\mathcal L_{\rm WTA}$, supervises the candidate
closest to the logged trajectory. Joint training combines all four losses:
\begin{equation}
\mathcal L = \mathcal L_{\rm WTA}+\mathcal L_{\rm score}
+\lambda_{\rm fut}\mathcal L_{\rm fut}
+\lambda_{\rm bank}\mathcal L_{\rm bank},
\label{eq:training-objective}
\end{equation}
with $\lambda_{\rm fut}=1$ and $\lambda_{\rm bank}=0.5$.
We down-weight the bank loss to limit its influence on shared features while
retaining supervision from trajectories outside the generated candidate set.

\Needspace{9\baselineskip}
\subsection{Inertial Re-Ranking}
\label{sec:method-state}

We observe abrupt changes between plans selected in adjacent frames.
Framewise scoring can favor a high-scoring trajectory that is incompatible
with the previous choice; similar candidate scores can also make selection
sensitive to small changes. We therefore check each candidate against the previous plan $\pi_{t-1}$ and use this compatibility to re-rank the current candidates.

\noindent
\begin{minipage}[t]{0.60\columnwidth}
\vspace{0pt}
\setlength{\emergencystretch}{1em}
For NAVSIM-v2, we instantiate the check with the benchmark's two-frame
extended-comfort (EC) criterion~\citep{navsimv2}. A deterministic
linear-quadratic-regulator (LQR) controller tracks the previous plan and each
current candidate with a kinematic bicycle model, producing ego-state rollouts.
After aligning the two rollouts on their common horizon, we compare their
acceleration, jerk, yaw-rate, and yaw-acceleration profiles. The candidate is
marked compatible when all four RMS differences fall below their respective
benchmark thresholds, giving the binary flag
\[
 c_{t,i}=c_{\rm EC}(\pi_{t-1},\tau_{t,i})\in\{0,1\}.
\]
Logged human plans pass this check on \humanEcPass{}\% of adjacent frame
pairs (Appendix~\ref{sec:appendix-ec-taxonomy}).
\end{minipage}\hfill
\begin{minipage}[t]{0.385\columnwidth}
\vspace{0pt}
\centering
\includegraphics[width=\linewidth]{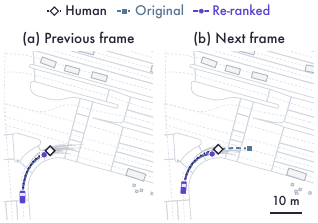}
\captionsetup{font=small,skip=3pt,justification=RaggedRight,
singlelinecheck=false,width=\linewidth}
\captionof{figure}{\textbf{Consecutive plans through a right turn.}
Frames are $0.5$\,s apart. In (b), re-ranking selects a plan closer to the
preceding choice and the human trajectory. Both views share map coordinates.}
\label{fig:ec-traj}
\end{minipage}
\par\smallskip\noindent
The check uses only the previous plan, current candidates, and controller
rollouts, so it requires no future observations or labels at deployment.
Within each log, we process frames chronologically and carry the selected
plan forward. The first frame uses the learned score alone.

We add the compatibility result as a log-domain penalty to the learned utility:
\begin{equation}
\Psi_{t,i}\coloneqq\log(0.01+0.99\,c_{t,i}),\qquad
 i_t^\star\in\arg\max_{1\le i\le N}
 \bigl[\hat J_{t,i}+\lambda_{\rm state}\Psi_{t,i}\bigr].
\label{eq:stateful-selection}
\end{equation}
With $\lambda_{\rm state}=1$, compatible candidates receive no penalty and
retain their learned-score ordering; incompatible candidates receive
$\Psi_{t,i}=-\log 100$. Re-ranking adds no learned parameters or training loss.
We report both framewise and re-ranked results on NAVSIM-v2.

\section{Experiments}
\label{sec:experiments}

We organize the experiments around four research questions:
\begingroup
\normalsize\rmfamily
\begin{list}{}{%
  \setlength{\leftmargin}{1.15em}%
  \setlength{\labelwidth}{0.55em}%
  \setlength{\labelsep}{0.60em}%
  \setlength{\topsep}{4pt}%
  \setlength{\itemsep}{2pt}%
  \setlength{\parsep}{0pt}%
}
\item[\textcolor{RQpurple}{$\bullet$}]
\textbf{RQ1.}\enspace\emph{Does World4Scorer improve driving performance on NAVSIM and Bench2Drive?}
\item[\textcolor{RQblue}{\small$\blacksquare$}]
\textbf{RQ2.}\enspace\emph{What do future targets, candidate outcomes, and inertial re-ranking contribute?}
\item[\textcolor{RQgreen}{\small$\blacklozenge$}]
\textbf{RQ3.}\enspace\emph{Do candidate-specific predictive states guide the model's trajectory selection?}
\item[\textcolor{RQamber}{\small$\blacktriangle$}]
\textbf{RQ4.}\enspace\emph{Can outcome-based scoring improve manipulation with a frozen world model?}
\end{list}
\endgroup
\newcommand{\venuecite}[2]{\hyperlink{cite.#2}{#1}\nocite{#2}}
\newcommand{\benchmarkcontexttable}{%
\begin{table*}[t]
\centering
\fontsize{6.4}{7.7}\selectfont
\setlength{\tabcolsep}{3.01pt}
\renewcommand{\arraystretch}{1.18}
\begin{tabular}{lll
 *{12}{>{\color{SubTone}}r}
 rr@{}}
\toprule
Method & Venue & Input & NC$^{2}$ & DAC$^{2}$ & TLC & DDC & EP$^{1}$ & EP$^{2}$ & TTC$^{1}$
 & TTC$^{2}$ & C & LK & HC & EC & PDMS & EPDMS \\
\addlinespace[1.6pt]
\midrule
\rowcolor{TableDirect}[\tabcolsep][\tabcolsep]\multicolumn{17}{@{}l}{\textbf{End-to-end}} \\
TransFuser & \venuecite{TPAMI'23}{transfuser} & C+L & \mrgTransFuser \\
\rowcolor{TableZebra}[\tabcolsep][\tabcolsep]DiffusionDrive & \venuecite{CVPR'25}{diffusiondrive} & C+L & \mrgDiffusionDrive \\
DiffusionDrive-V2 & \venuecite{arXiv'25}{diffusiondrivev2} & C & \mrgDiffusionDriveVTwo \\
\rowcolor{TableZebra}[\tabcolsep][\tabcolsep]DAWN & \venuecite{arXiv'26}{dawn} & C & \mrgDAWN \\
LWDrive & \venuecite{arXiv'26}{lwdrive} & C & \mrgLWDrive \\
\rowcolor{TableZebra}[\tabcolsep][\tabcolsep]UniTeD & \venuecite{ECCV'26}{united} & C & \mrgUniTeD \\
DrivoR & \venuecite{CVPR'26}{drivor} & C & \mrgDrivorReleased \\
\addlinespace[1.6pt]
\midrule
\rowcolor{TableVLA}[\tabcolsep][\tabcolsep]\multicolumn{17}{@{}l}{\textbf{Vision-language-action}} \\
ReCogDrive & \venuecite{ICLR'26}{recogdrive} & C & \mrgReCogDrive \\
\rowcolor{TableZebra}[\tabcolsep][\tabcolsep]DriveVLA-W0 & \venuecite{ICLR'26}{drivevlaw0} & C & \mrgDriveVLAWZero \\
DriveWorld-VLA & \venuecite{ICML'26}{driveworldvla} & C & \mrgDriveWorldVLA \\
\rowcolor{TableZebra}[\tabcolsep][\tabcolsep]ExploreVLA & \venuecite{ECCV'26}{explorevla} & C & \mrgExploreVLA \\
WAM-Flow & \venuecite{CVPR'26}{wamflow} & C & \mrgWAMFlow \\
\addlinespace[1.6pt]
\midrule
\rowcolor{TableWorld}[\tabcolsep][\tabcolsep]\multicolumn{17}{@{}l}{\textbf{World-model augmented}} \\
WoTE & \venuecite{ICCV'25}{wote} & C+L & \mrgWoTE \\
\rowcolor{TableZebra}[\tabcolsep][\tabcolsep]WorldRFT & \venuecite{AAAI'26}{worldrft} & C & \mrgWorldRFT \\
GraphWorld & \venuecite{arXiv'26}{graphworld} & C & \mrgGraphWorld \\
\rowcolor{TableZebra}[\tabcolsep][\tabcolsep]Discrete-WAM & \venuecite{arXiv'26}{discretewam} & C & \mrgDiscreteWAM \\
DriveFuture & \venuecite{arXiv'26}{drivefuture} & C & \mrgDriveFuture \\
\rowcolor{TableZebra}[\tabcolsep][\tabcolsep]SafeDrive & \venuecite{CVPR'26}{safedrive} & C+L & \mrgSafeDrive \\
Drive-JEPA & \venuecite{arXiv'26}{drivejepa} & C & \mrgDriveJEPA \\
\addlinespace[1.6pt]
\midrule
\rowcolor{TableOurs}[\tabcolsep][\tabcolsep]\textbf{Ours} & This work & C & \mrgOursOff \\
\rowcolor{TableOurs}[\tabcolsep][\tabcolsep]\quad $+$ inertial re-ranking & & C & \mrgOursOn \\
\bottomrule
\end{tabular}
\caption{\textbf{Planning performance on NAVSIM-v1 and NAVSIM-v2 (\%).}
Input C: cameras, L: LiDAR; superscripts 1 and 2 denote native NAVSIM-v1 and NAVSIM-v2.
Best and second-best results are bold and underlined.
We evaluate with the latest official NAVSIM devkits.}
\label{tab:navsim-v1-context}
\label{tab:navsim-v2-context}
\end{table*}
}

\benchmarkcontexttable
\begin{table}[t]
\centering
\small
\begingroup
\fontsize{8.5}{10}\selectfont
\setlength{\tabcolsep}{3.5pt}
\renewcommand{\panelgap}{11pt}
\renewcommand{\arraystretch}{1.00}
\setlength{\aboverulesep}{1.5pt}
\setlength{\belowrulesep}{1.5pt}
\begin{tabular}{ccr p{\dimexpr\panelgap-2\tabcolsep\relax}
                ccr p{\dimexpr\panelgap-2\tabcolsep\relax} ccr}
\multicolumn{3}{c}{\textbf{(a) Auxiliary target}} & &
\multicolumn{3}{c}{\textbf{(b) Outcome-label pairing}} & &
\multicolumn{3}{c}{\textbf{(c) Inertial re-ranking}} \\

\cmidrule{1-3}\cmidrule{5-7}\cmidrule{9-11}
Aux. & Realized & PDMS\hphantom{\tabdelta{\pdmsLeanjfullLast}{\pdmsLeanjzeroLast}} & &
Permuted & Sel. & Order & &
Score & off & on\hphantom{\tabdelta{\vTwoStateUnifiedNavTwoOne}{\vTwoRawUnifiedNavTwoOne}} \\
\cmidrule{1-3}\cmidrule{5-7}\cmidrule{9-11}
\xmark & \xmark & \pdmsLeanjzeroLast\hphantom{\tabdelta{\pdmsLeanjfullLast}{\pdmsLeanjzeroLast}} & &
none & \permPdmControl & \permOrderControl & &
EC & \ecUnifiedNavTwoOffOne & \ecUnifiedNavTwoOnOne\tabdelta{\ecUnifiedNavTwoOnOne}{\ecUnifiedNavTwoOffOne} \\
\cmark & \xmark & \pdmsLeanjcurLast\tabdelta{\pdmsLeanjcurLast}{\pdmsLeanjzeroLast} & &
candidate bank & \permPdmClover & \permOrderClover & &
EP & \epUnifiedNavTwoOffOne & \epUnifiedNavTwoOnOne\tabdelta{\epUnifiedNavTwoOnOne}{\epUnifiedNavTwoOffOne} \\
\cmark & \cmark &
\pdmsLeanjfullLast\tabdelta{\pdmsLeanjfullLast}{\pdmsLeanjzeroLast} & &
all & \permPdmAll & \permOrderAll & &
EPDMS & \vTwoRawUnifiedNavTwoOne
& \vTwoStateUnifiedNavTwoOne\tabdelta{\vTwoStateUnifiedNavTwoOne}{\vTwoRawUnifiedNavTwoOne} \\
\cmidrule[\heavyrulewidth]{1-3}\cmidrule[\heavyrulewidth]{5-7}\cmidrule[\heavyrulewidth]{9-11}
\end{tabular}
\endgroup
\caption{\textbf{Ablations of future targets, outcome supervision, and re-ranking.}
(a)~Native-V1 PDMS with no, current-frame, or realized-future
targets. (b)~Outcome-label permutation during training, tested on
\permEvalScenes{} held-out scenes. \emph{Sel.}: selected PDM
(0--1; random pick: \permRandomPick{}). \emph{Order}: pairwise ordering
accuracy. (c)~Native-V2 re-ranking with fixed weights and candidates.
EC: extended comfort; EP: ego progress.
Subscripts give changes from row 1 in (a) and off in (c).}
\label{tab:rq2-target}
\label{tab:rq4-transfer}
\label{tab:rq3-transfer}

\end{table}

\Needspace{8\baselineskip}
\subsection{Setup}

We evaluate on all \navtestTokens{} NAVSIM test scenes using the NAVSIM-v1
PDM score (PDMS)~\citep{navsim} and NAVSIM-v2 extended PDMS
(EPDMS)~\citep{navsimv2}, defined in Appendix~\ref{sec:appendix-score-computation}.
Both metrics score the selected trajectory in the official simulator;
the learned utility only ranks candidates.
World4Scorer uses DrivoR's~\citep{drivor} training split,
LoRA-adapted DINOv2 encoder, generator, and candidate count;
we train for 25 epochs with AdamW, a peak learning rate of
$2\times10^{-4}$, and a global batch size of 64. Linear warmup covers the
first 10\% of training, then cosine decay. We select the main checkpoint
on the validation split.

\begin{figure}[t]
\centering
\includegraphics[width=\linewidth]{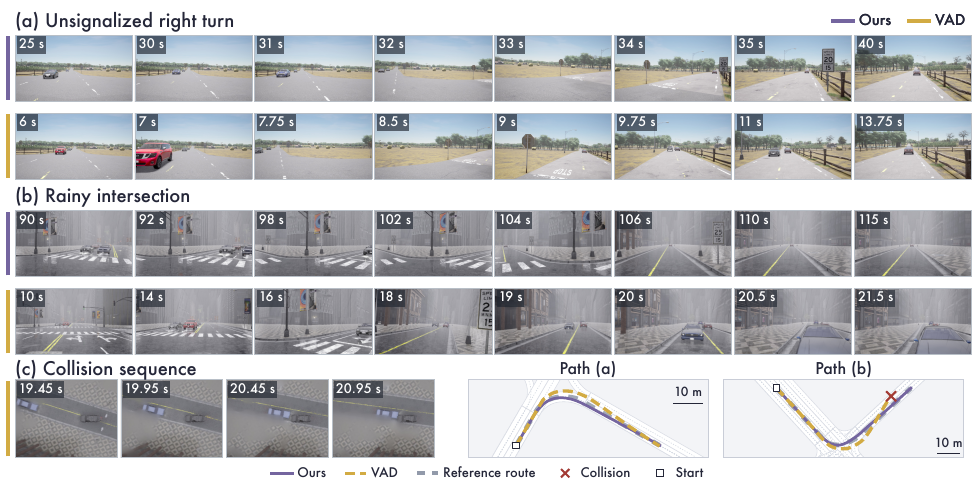}
\captionsetup{font=small,skip=3pt,justification=RaggedRight,singlelinecheck=false}
\caption{\textbf{Closed-loop trajectories on Bench2Drive.}
Purple denotes Ours; yellow denotes VAD~\citep{vad}.
Eight frames per method show (a)~a right turn where VAD leaves its route lane
and (b)~a rainy intersection where VAD collides with an oncoming vehicle.
Ours completes both routes. Panel (c) enlarges the collision sequence;
maps show the full executed paths and recorded collision location.}
\label{fig:b2d-main}
\end{figure}

\subsection{\texorpdfstring{\rqbadge{RQpurple}{$\bullet$}}{}Driving Performance (RQ1)}
\label{sec:rq-effectiveness}

\textbf{Planning on NAVSIM.}
World4Scorer achieves state-of-the-art NAVSIM-v2 performance with
\vTwoStateUnifiedNavTwoOne{} EPDMS and obtains
\pdmsUnifiedFullBestMain{} PDMS on NAVSIM-v1, the best results in
Table~\ref{tab:navsim-v1-context}.
Without re-ranking, it already reaches
\vTwoRawUnifiedNavTwoOne{} EPDMS, \prosedelta{\vTwoRawUnifiedNavTwoOne}{\epdmsReleasedSameHostOne}{} points above
DrivoR's \epdmsReleasedSameHostOne{} on the same evaluation. Re-ranking adds \prosedelta{\vTwoStateUnifiedNavTwoOne}{\vTwoRawUnifiedNavTwoOne}{} EPDMS points
through previous-plan compatibility.
The comparison evaluates complete planners with their own
generators; Section~\ref{sec:rq-mechanism} tests scoring with fixed candidate pools.

\begin{wraptable}[10]{r}{0.44\columnwidth}
\vspace{-11pt}
\centering
\fontsize{8.5}{10}\selectfont
\setlength{\tabcolsep}{3pt}
\renewcommand{\arraystretch}{1.03}
\begin{tabular*}{\linewidth}{@{\extracolsep{\fill}}lr@{}}
\toprule
Method & Score $\uparrow$ \\
\midrule
VAD~\citep{vad} & 42.35 \\
WoTE~\citep{wote} & 61.71 \\
ThinkTwice~\citep{thinktwice} & 62.44 \\
DriveAdapter~\citep{driveadapter} & 64.22 \\
Drive-JEPA~\citep{drivejepa} & 64.52 \\
SafeDrive~\citep{safedrive} & \underline{66.77} \\
\textbf{Ours} & \textbf{\bTwoDOurs} \\
\bottomrule
\end{tabular*}
\captionsetup{font=small,skip=3pt,width=\linewidth,justification=RaggedRight,singlelinecheck=false}
\caption{\textbf{Bench2Drive results.} Driving Score.}
\label{tab:b2d-main}
\end{wraptable}

\noindent
\textbf{Closed-loop driving on Bench2Drive.}
After training on the official 1000-clip subset, our adapted planner achieves
\bTwoDOurs{} Driving Score and \bTwoDSR{}\% success over 220
Bench2Drive~\citep{bench2drive} routes (Table~\ref{tab:b2d-main}).
The adaptation adds a route-point input and uses simulator ego state,
retained turn commands, and route-based re-ranking (Appendix~\ref{sec:b2d-closed-loop}). The table compares this
complete system with published reference results; their training data and
input interfaces are not matched to ours.
Figure~\ref{fig:b2d-main} compares two recorded maneuvers with VAD~\citep{vad}.
In (a), VAD completes the right turn but leaves its route lane, scoring 77.77.
In (b), it enters the opposing lane after the turn, collides with an oncoming
vehicle, and becomes blocked, scoring 43.93.
Ours completes both routes without lane departures or collisions,
achieving a Driving Score of 100 in each case.

\subsection{\texorpdfstring{\rqbadge{RQblue}{$\blacksquare$}}{}Supervision and Re-Ranking (RQ2)}
\label{sec:rq-components}

\textbf{Realized-future targets.}
Anchoring the executed state to its observed future raises PDMS from
\pdmsLeanjzeroLast{} to \pdmsLeanjfullLast{}, whereas the same objective with
a current-frame target gives \pdmsLeanjcurLast{} (Table~\ref{tab:rq2-target}a).
The three runs share the architecture, seed, bank supervision, and last-epoch
evaluation; this single-seed gap is consistent with, but does not by itself
isolate, the targets' temporal content.

\noindent
\textbf{Candidate--outcome pairing.}
Shuffling bank outcome labels reduces ordering accuracy from \permOrderControl{}
to \permOrderClover{}; shuffling all labels gives \permOrderAll{}
(Table~\ref{tab:rq2-target}b). Selected PDM falls from \permPdmControl{}
to \permPdmClover{} and \permPdmAll{}. Each permutation preserves the
scene's trajectories and label distribution but breaks their correspondence.
All three conditions train the same architecture on 512 scenes for 60 epochs
and evaluate on \permEvalScenes{} scenes from disjoint logs. These controls
show that ranking requires the correct correspondence between trajectories and outcomes.

\noindent
\textbf{Inertial re-ranking.}
\label{sec:rq-stabilization}
With weights and candidates fixed, inertial re-ranking raises extended
comfort from \ecUnifiedNavTwoOffOne{} to \ecUnifiedNavTwoOnOne{} while
progress changes only from \epUnifiedNavTwoOffOne{} to
\epUnifiedNavTwoOnOne{}, yielding \prosedelta{\vTwoStateUnifiedNavTwoOne}{\vTwoRawUnifiedNavTwoOne}{} EPDMS points
(Table~\ref{tab:rq4-transfer}c). In the right turn of Figure~\ref{fig:ec-traj},
it favors trajectories compatible with the preceding choice.

\begin{figure}[!t]
\centering
\includegraphics[width=\textwidth]{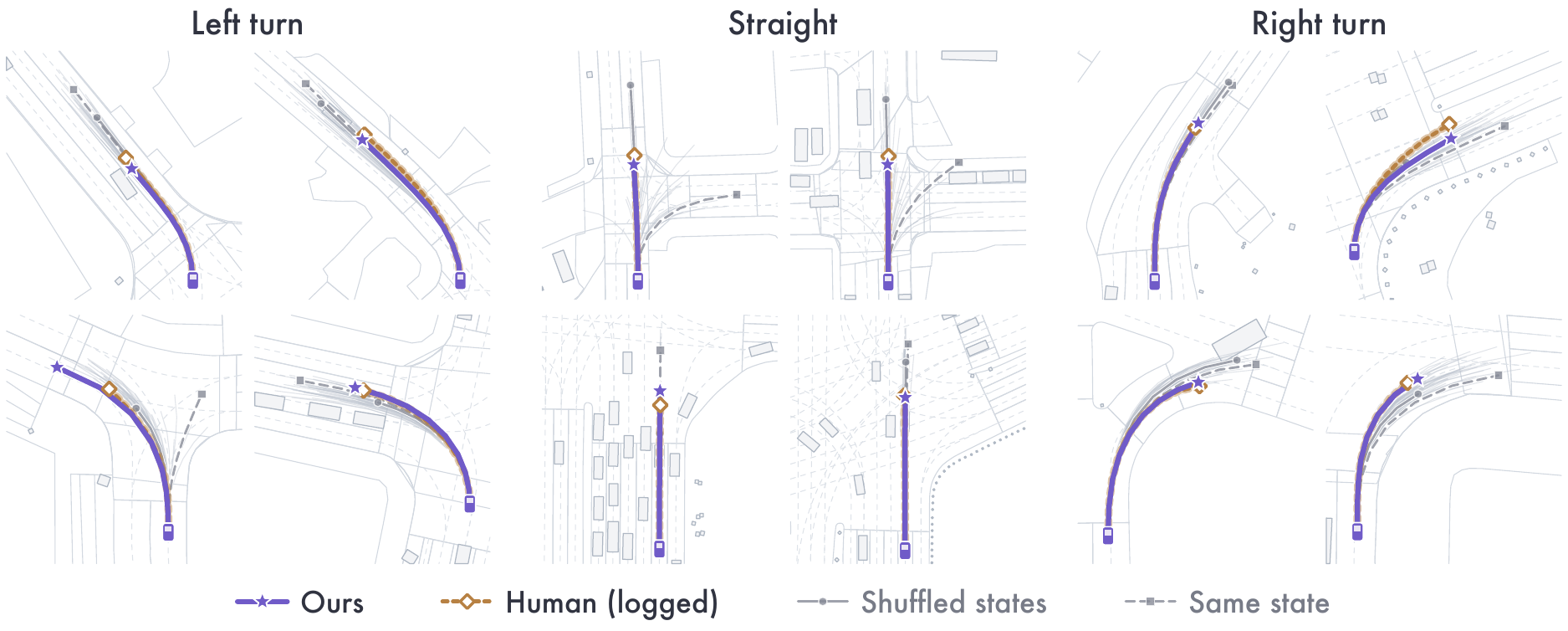}
\captionsetup{font=small,skip=3pt,justification=RaggedRight,singlelinecheck=false}
\caption{\textbf{Candidate-specific states guide trajectory selection.}
Twelve selected scenes where \emph{Ours} attains pool-best native-V1 PDMS.
Purple and amber show Ours and the logged human trajectory; gray marks state
interventions. Faint lines show the same 64-candidate pool used in all three conditions.}
\label{fig:representation}
\end{figure}

\begin{figure}[!t]
\centering
\includegraphics[width=\textwidth]{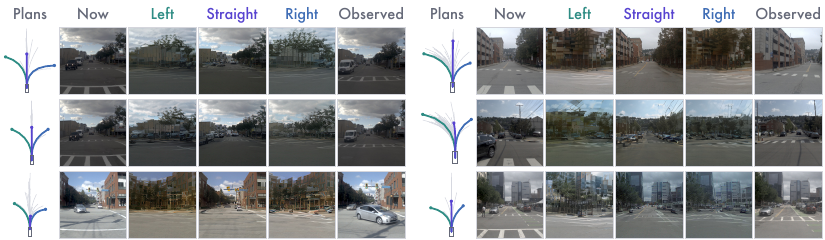}
\captionsetup{font=small,skip=3pt,justification=RaggedRight,singlelinecheck=false}
\caption{\textbf{Visual readouts for alternative trajectory queries.}
Colors match left-turn, near-straight (selected), and right-turn candidates
to their $+2$\,s readouts from a separately trained network.
Only the logged trajectory has an observed future; query images are
qualitative readouts, not tests of counterfactual accuracy.}
\label{fig:latent-query-scan}
\end{figure}

\Needspace{4\baselineskip}
\subsection{\texorpdfstring{\rqbadge{RQgreen}{$\blacklozenge$}}{}Trajectory-Selection Analysis (RQ3)}
\label{sec:rq-mechanism}
\label{sec:rq-target-source}

\textbf{Matching states to candidates.}
We test whether selection depends on the state predicted for each trajectory.
The model, scene features, and \figTwoNumCandidates{} candidates per scene
stay fixed; we change only the assignment of states to candidates.
Every evaluation covers all \navtestSize{} NAVSIM-v1 test scenes.

\begin{wraptable}[10]{r}{0.47\columnwidth}
\vspace{-5pt}
\centering
\small
\setlength{\tabcolsep}{4pt}
\renewcommand{\arraystretch}{1.12}
\begin{tabular}{p{\dimexpr0.54\linewidth-2\tabcolsep\relax}
 >{\raggedleft\arraybackslash}p{\dimexpr0.25\linewidth-2\tabcolsep\relax}
 >{\raggedleft\arraybackslash}p{\dimexpr0.21\linewidth-2\tabcolsep\relax}}
\toprule
Candidate state & PDMS $\uparrow$ & Drop $\downarrow$ \\
\midrule
\textbf{Own state} & \textbf{\rSixMatchedSelected{}} & -- \\
Shuffled states & \rSixRepeatSelected{} & \rSixRepeatDrop{} \\
Scene-mean state & \rSixMeanSelected{} & \prosedelta{\rSixMatchedSelected}{\rSixMeanSelected}{} \\
\bottomrule
\end{tabular}
\captionsetup{font=small,skip=3pt,width=\linewidth,justification=RaggedRight,singlelinecheck=false}
\caption{\textbf{Candidate states affect selection.}
All \navtestSize{} test scenes with DrivoR's 64 candidates, not the Table~\ref{tab:navsim-v1-context} pool.
Shuffled PDMS averages 100 runs; drops are relative to Own state.}
\label{tab:candidate-state-selection}
\end{wraptable}

\noindent
Assigning each trajectory another candidate's state reduces PDMS from
\rSixMatchedSelected{} to \rSixRepeatSelected{}, averaged over 100 random shuffles
(Table~\ref{tab:candidate-state-selection}). Replacing all states with their
scene mean yields \rSixMeanSelected{}. Equal scores then leave the choice
to a fixed candidate index.

Every one of the 100 shuffles lowers PDMS.
Shuffling preserves each scene's score distribution but breaks the match
between trajectories and states, changing which candidate receives each score.

Figure~\ref{fig:representation} illustrates twelve scenes where original states
select pool-best plans and reassignment changes the choice.
Successful plans can differ from the logged driver because the evaluator
scores safety, progress, and comfort along each candidate.

\noindent
\textbf{Visualizing candidate states.}
Figure~\ref{fig:latent-query-scan} compares left-turn, near-straight, and
right-turn queries in each scene.
With the driving model frozen, we train a separate image reconstruction
network on \decodeTrainPairs{} logged current/future pairs. It combines
predicted pooled features with the current spatial grid and a frozen
RAE~\citep{rae} decoder. Query changes alter the decoded viewpoint. These images visualize query-conditioned representations;
only the logged query has an observed future.

\begin{figure}[!t]
\centering
\includegraphics[width=\columnwidth]{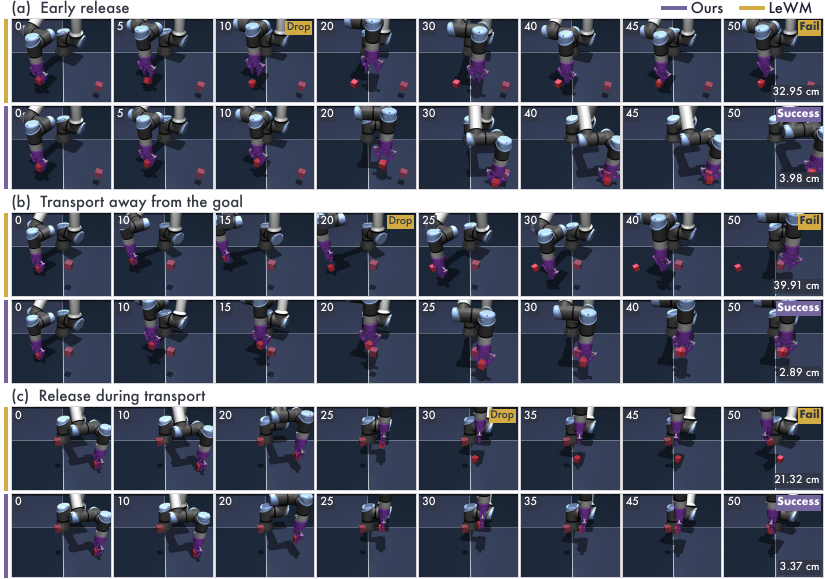}
\captionsetup{font=small,skip=2pt,justification=RaggedRight,singlelinecheck=false,width=\linewidth}
\caption{\textbf{Outcome-based scoring corrects unsuccessful LeWM plans.}
For each case, eight frames trace each method from the same start toward the same goal. Numbers mark steps;
\emph{Drop} marks release. Badges give outcomes and final distances to the
translucent goal cube (success: within \cubeToleranceCm{}\,cm).}
\label{fig:cube-pair}
\end{figure}

\Needspace{9\baselineskip}
\subsection{\texorpdfstring{\rqbadge{RQamber}{$\blacktriangle$}}{}Transfer to Manipulation (RQ4)}
\label{sec:rq-transfer}

\begin{wraptable}[12]{r}{0.40\columnwidth}
\vspace{-11pt}
\centering
\fontsize{8.5}{10}\selectfont
\setlength{\tabcolsep}{2pt}
\renewcommand{\arraystretch}{1.02}
\newcommand{\cubemethod}[2]{#1\enspace{\fontsize{7}{8}\selectfont\citep{#2}}}
\begin{tabular*}{\linewidth}{@{\extracolsep{\fill}}lr@{}}
\toprule
Method & Success (\%) \\
\midrule
\cubemethod{GCIQL}{iql} & \cubeGCIQL \\
\cubemethod{GCIVL}{ogbench} & \cubeGCIVL \\
\cubemethod{PLDM}{pldm} & \cubePLDM \\
\midrule
\cubemethod{LeWM}{lewm} & \cubeAnchorTenMean \\
\textbf{Ours} & \textbf{\cubeOursTenMean} \\
\bottomrule
\end{tabular*}
\captionsetup{font=small,skip=3pt,width=\linewidth,justification=RaggedRight,singlelinecheck=false}
\caption{\textbf{OGBench-Cube results.} Upper rows are reported by LeWM;
lower rows average our \cubeReportSeeds{} seeds of \cubeEpisodes{} paired episodes.}
\label{tab:cube-main}
\end{wraptable}

On OGBench-Cube~\citep{ogbench}, adding outcome-based scoring to
LeWM~\citep{lewm} raises success from \cubeAnchorTenMean{}\% to
\cubeOursTenMean{}\% over \cubeReportSeeds{} evaluation seeds (Table~\ref{tab:cube-main}; paired gain
\cubeDeltaTen{} points, 95\% CI \cubeDeltaTenCI{}). We keep LeWM's official
implementation, world model, and planning budget fixed; our scorer evaluates
action sequences from predicted states and goals and adds its learned outcome
score to LeWM's planning score.

Each seed runs \cubeEpisodes{} paired episodes sharing initial states and goals
(Appendix~\ref{sec:appendix-cube}). At LeWM's released evaluation seed
\cubeSeedMain{}, success rises from \cubeAnchorTenSeedMain{}\% to \cubeOursTenSeedMain{}\%.
In Figure~\ref{fig:cube-pair}, LeWM ends 21--40\,cm from the goal
and ours succeeds.

\FloatBarrier %
\section{Conclusion}
\label{sec:conclusion}

World4Scorer scores candidates with a trajectory-conditioned predictor trained
by simulator outcomes and anchored by the observed future. Controlled
experiments show the value of correctly paired outcome labels and
realized-future targets, and state interventions show that selection depends on
candidate-specific states. The method achieves state-of-the-art NAVSIM-v2
performance and strong closed-loop driving results, and outcome-based scoring
also improves manipulation with a frozen world model. These findings support
grounding predictive states in decision outcomes.

\FloatBarrier
\clearpage

\section*{Reproducibility Statement}
All experiments use public benchmarks and their official evaluators
(NAVSIM-v1 and NAVSIM-v2, Bench2Drive, and OGBench-Cube).
Appendix~\ref{sec:appendix-training-spec} and
Table~\ref{tab:appendix-training-config} give the full training configuration,
including the data split, losses and weights, optimizer, schedule, batch size,
and checkpoint selection; Appendix~\ref{sec:appendix-compute} reports model
size, compute, and latency. Appendix~\ref{sec:appendix-score-computation}
specifies how the simulator labels used for training are computed, and
Section~\ref{sec:method-state} with Appendix~\ref{sec:appendix-training-spec}
specifies inertial re-ranking, including the controller and thresholds. The
fixed-pool interventions (Appendix~\ref{sec:appendix-pool-audit}) state their
permutation seeds and bootstrap procedures, and Appendices~\ref{sec:appendix-cube}
and~\ref{sec:b2d-closed-loop} describe the OGBench-Cube and Bench2Drive
protocols. We will release the code upon publication.

\section*{Use of AI}
Large language models assisted with prose editing, LaTeX, figure code,
analysis scripts, and cluster tooling. The authors determined the scientific
claims, experimental design, controls, and interpretation of the results.
Reported results come from experiments and cited studies. The authors verified
the manuscript and take full responsibility for its content.

\begingroup
\setlength{\emergencystretch}{1em}
\bibliography{references}
\bibliographystyle{iclr2027_conference}
\endgroup

\clearpage
\appendix
\flushbottom
\newcommand{\appendixcolhead}[1]{#1}
\newcommand{\appendixfinding}[2]{\paragraph{#1}#2\par}

\section*{Appendix}

This appendix provides implementation details and further analyses of prediction and planning.

\begin{itemize}
\item Appendix~\ref{sec:appendix-diagnostics}: notation, training objectives,
implementation, and computational cost.
\item Appendix~\ref{sec:appendix-evaluation}: simulator labels and differences
between NAVSIM scoring rules.
\item Appendix~\ref{sec:appendix-representations}: visual targets, image
readouts, and candidate-dependent features.
\item Appendix~\ref{sec:appendix-selection}: state reassignment and
transfer to manipulation.
\item Appendix~\ref{sec:appendix-sequential}: consecutive selections,
history controls, and Bench2Drive evaluation.
\end{itemize}

\needspace{6\baselineskip}
\section{Setup and Reference}
\label{sec:appendix-diagnostics}
We define the notation and document World4Scorer's training setup and computational cost.
\subsection{Notation Reference}
\label{sec:appendix-notation}

\begin{table}[H]
\centering
\begin{minipage}{\columnwidth}
\centering
\small
\setlength{\tabcolsep}{5pt}
\renewcommand{\arraystretch}{1.02}
\begin{tabular*}{\linewidth}{@{}p{0.29\linewidth}@{\extracolsep{\fill}}p{0.67\linewidth}@{}}
\toprule
\appendixcolhead{Symbol} & \appendixcolhead{Meaning} \\
\midrule
$t$, $i,j$, $N$ & Decision time, candidate indices, and candidate count ($N=64$). \\
$n$, $M$ & Training-scene index and scene count in the future loss. \\
$o_t$, $o_{t+2\mathrm{s}}$ & Current observation and the observed future two seconds later. \\
$s_t$, $e_t$ & Camera scene tokens and the encoded ego state. \\
$\tau_{t,i}$, $\mathcal C_t$, $\tau_t^{\rm exec}$ & Candidate trajectory, candidate set, and logged executed trajectory. \\
$\phi$, $P$, $g$, $H$ & Action encoder, shared predictor, visual readout, and score heads. \\
$q_{t,i}$, $Q_t$ & Candidate query and the collection of candidate queries. \\
$z_{t,i}$, $Z_t$, $z_t^{\rm exec}$ & Candidate state, candidate-state matrix, and executed-query state. \\
$f_{\rm DINO}$, $u_t^{\rm fut}$ & Frozen visual encoder and its observed future embedding. \\
$\operatorname{sg}$, $\sigma$ & Stop-gradient operator and sigmoid function. \\
$\ell_{t,i,m}$, $p_{t,i,m}$, $y_m$ & Component logit, sigmoid prediction, and outcome target. \\
$\mathcal G$, $\mathcal A$, $\mathcal M$ & Multiplicative and additive score-component sets; $\mathcal M=\mathcal G\cup\mathcal A$. \\
$\hat J_{t,i}$, $\alpha_m$, $\beta_m$ & Log-domain ranking score and its fixed component weights. \\
$\mathcal L_{\rm WTA}$, $\mathcal L_{\rm score}$ & Imitation loss and generated-candidate outcome loss. \\
$\mathcal L_{\rm fut}$, $\mathcal L_{\rm bank}$ & Observed-future loss and candidate-bank outcome loss. \\
$\mathcal L_{\rm out}$, $\ell_{\rm out}$, $w_i$ & Outcome objective, per-candidate loss, and averaging/weighting factor. \\
$\lambda_{\rm fut}$, $\lambda_{\rm bank}$ & Future and bank loss weights ($1$ and $0.5$). \\
$\pi_{t-1}$, $c_{t,i}$, $c_{\rm EC}$ & Previous selected plan, binary compatibility result, and the EC check. \\
$\Psi_{t,i}$, $\lambda_{\rm state}$, $i_t^\star$ & Re-ranking penalty, its weight ($1$), and the selected index. \\
$\theta_0$, $\delta$ & Fitted parameters of $\phi,P,g,H$ and a local perturbation. \\
$A_F$, $A_O$ & Weighted Jacobians of normalized visual and outcome predictions. \\
$G_O$, $G$ & Local constraint matrices before and after adding visual constraints. \\
$\hat\Delta_{ij}$, $c_{ij}$ & Candidate score difference and its parameter gradient at $\theta_0$. \\
$\mu$, $\lambda$, $\epsilon$ & Analysis scale, visual-constraint weight, and perturbation budget. \\
$\rho_{ij}(G)$ & Maximal first-order score-difference change; $\rho_{ij}^{\rm out}=\rho_{ij}(G_O)$, $\rho_{ij}^{\rm out+fut}=\rho_{ij}(G)$. \\
$\widehat u_{t,i}$, $X_t$, $\widehat X_{t+2\mathrm{s},i}$ & Predicted visual feature, current patch grid, and reconstructed future grid. \\
$\lambda_{\rm cube}$ & Weight of learned utility in the LeWM planner. \\
PDMS, EPDMS & Native NAVSIM-v1 and NAVSIM-v2 aggregate planning scores. \\
\bottomrule
\end{tabular*}
\captionsetup{width=\linewidth,justification=RaggedRight,singlelinecheck=false}
\caption{\textbf{Notation used in the method and analyses.}
Time indices are omitted when a scene is fixed.
Section~\ref{sec:method-base} defines the six predicted score components.}
\label{tab:notation}
\end{minipage}
\end{table}

\subsection{Training and State Interventions}
\label{sec:appendix-training-spec}

\paragraph{Trajectory and outcome losses.}
Generated, executed, and bank queries share the predictor $P$. The heads $H$
score candidate states, while $g$ reads the executed state for visual prediction.
For generated candidates, $\mathcal L_{\rm WTA}$ sums
absolute errors over coordinates, averages over poses, minimizes over
candidates, and then averages over scenes. We sum this loss for two targets:
the logged four-second trajectory and the same trajectory time-warped to
five seconds by cubic-spline resampling.

The outcome losses average over scene--candidate pairs, written
$\langle\cdot\rangle_{\rm gen}$ and $\langle\cdot\rangle_{\rm bank}$:
\begin{equation}
\begin{aligned}
\mathcal L_{\rm score}
&=\sum_{m\in\mathcal M}\langle b(\ell_m,y_m)\rangle_{\rm gen},\\
\mathcal L_{\rm bank}
&=\frac15\sum_{m\in\mathcal M\setminus\{\mathrm{EP}\}}
\langle b(\ell_m,y_m)\rangle_{\rm bank}
+0.1\langle|\sigma(\ell_{\rm EP})-y_{\rm EP}|\rangle_{\rm bank}.
\end{aligned}
\label{eq:component-losses}
\end{equation}
Here $\mathcal M=\mathcal G\cup\mathcal A=\{\mathrm{NC,DAC,DDC,TTC,EP,C}\}$,
$y_m$ is the outcome target for component $m$, and
$b(\ell,y)=\log(1+\exp\ell)-y\ell$ is binary cross-entropy on a logit.
Generated progress uses soft-label cross-entropy, whereas bank progress uses
L1. NC and DDC targets equal to $0.5$ are mapped to zero before these losses.
The averages use the available labels and masks described below.
Candidate coordinates are detached; the scoring losses update $\phi$, $P$, $H$, and the scene features.

\paragraph{Visual supervision.}
Once per scene, the executed query predicts the frozen front-view
embedding two seconds later. We minimize cosine distance with weight
$\lambda_{\rm fut}=1$. Gradients update $\phi,P,g$; current-scene tokens are
detached on this auxiliary path.

With one query, self-attention has only one key and its softmax is constant.
For a singleton input $x$ and head dimension $d$,
\begin{equation}
\operatorname{Attn}(x)
=\operatorname{softmax}\!\left(\frac{(W_Qx)^\top(W_Kx)}{\sqrt d}\right)W_Vx
=W_Vx,
\qquad \nabla_{W_Q,W_K}\mathcal L_{\rm fut}=0.
\label{eq:singleton-attention}
\end{equation}
Thus the self-attention query and key projections receive no visual-loss
gradient: \trunkQKUnreached{} of the \trunkSharedParams{} shared parameters
(\trunkQKUnreachedShare{}). They are trained by the multi-candidate outcome
losses. Cross-attention over scene tokens is unaffected by this singleton property.

\paragraph{Candidate-bank supervision.}
Each step samples \setDrawPerStep{} extra trajectories from the matching scene,
with bank loss weight $\lambda_{\rm bank}=0.5$.
Each scene's bank holds \setValidMean{} valid entries on average, so one step sees on average \setSeenShare{} of a scene's stored pool. There are no bank entries for
\setAbsentScenes{} scenes; \setQuotaShortShare{} of scenes lack enough
trajectories in the required safety categories to fill the quota.

Validity masks exclude missing outcome labels from the losses.
Bank time-to-collision labels are used without masking; generated-candidate
labels use $2.0$ as the invalid-target marker. All available components
supervise the shared predictor used for candidate scoring at deployment.

\paragraph{Training configuration.}
Table~\ref{tab:appendix-training-config} lists training settings. Caches store
front-view embeddings two seconds ahead and scene-matched trajectories, labels, and validity masks.

\begin{table}[H]
\centering\small
\renewcommand{\arraystretch}{1.12}
\begin{tabular}{@{}p{.22\linewidth}p{.73\linewidth}@{}}
\toprule
\appendixcolhead{Setting} & \appendixcolhead{Value} \\
\midrule
Data and batches & $103{,}288$ examples; four devices, $16$ samples per device ($64$ total). \\
Optimization & AdamW, $(\beta_1,\beta_2)=(0.9,0.999)$, $\epsilon=10^{-8}$, weight decay $0.01$. \\
Schedule & $25$ epochs; peak learning rate $2\times10^{-4}$, linear warmup over the first $10\%$ of steps, then cosine decay to zero. \\
Image input & $672\times1148$ (height $\times$ width), ImageNet normalization; deterministic preprocessing without crop, flip, or color jitter. \\
Backbone & DINOv2 ViT-S/14 with rank-$32$ LoRA. \\
Predictor & Four decoder layers; feature width $256$, feed-forward width $1024$. \\
Random seed & 3 for Python, NumPy, PyTorch, the distributed sampler, and data workers. \\
\bottomrule
\end{tabular}
\caption{\textbf{Training configuration.} All runs use the public DrivoR software stack.}
\label{tab:appendix-training-config}
\end{table}

Validation runs every epoch. The reported NAVSIM checkpoint is epoch 23,
selected by validation score; the three target controls in
Table~\ref{tab:rq2-target}a use the last epoch. Appendix~\ref{sec:b2d-closed-loop} gives the Bench2Drive setup. The test split is excluded from
training, validation, and checkpoint selection.

\paragraph{Candidate-state interventions.}
We fix the epoch-23 checkpoint, scene features, candidate pool, score heads,
and score aggregation. Scene-mean replacement assigns the mean state to every
candidate in a scene. State shuffling uses a fixed-point-free permutation,
so each candidate receives another candidate's state while all inputs remain fixed.

\paragraph{Inertial re-ranking and evaluation.}
Frames are processed chronologically within each log, and the selected
trajectory is carried to the next frame. At the nominal $0.5$\,s frame
interval, the $0.1$\,s LQR rollouts of the previous plan and each candidate
are compared over their shared $3.5$\,s window.
The first frame uses framewise selection. Ties are resolved by selecting the lowest candidate index.

Each selected trajectory is evaluated with the official benchmark aggregator;
selection uses the learned utility in Equation~\ref{eq:ranking-utility}.
The ordered component weights
$(\alpha_{\rm NC},\alpha_{\rm DAC},\alpha_{\rm DDC},
\beta_{\rm TTC},\beta_{\rm EP},\beta_{\rm C})$ are
$\selWeightsVOne$ for NAVSIM-v1 and $\selWeightsNavTwo$ for NAVSIM-v2.
Section~\ref{sec:rq-effectiveness} evaluates the epoch 23 checkpoint under
native benchmark scoring. Section~\ref{sec:rq-stabilization} toggles inertial
re-ranking while holding this checkpoint and its candidate pool fixed.

\paragraph{Compute.}
The 25-epoch training run uses four 80\,GB A800/A100 GPUs for roughly \trainHours{} hours.
On one 24\,GB A30, inference at batch 32 runs at \aThirtyRate{} frames per second with \aThirtyVram{}\,GiB peak memory
and processes the full test split in about \aThirtyFullHours{} hours. Software
follows the public DrivoR stack.

\subsection{Compute and Parameter Scale}
\label{sec:appendix-compute}

World4Scorer uses the DrivoR sensor stack. Table~\ref{tab:compute-scale}
compares the parameter counts, computation, and latency of planners that score multiple candidates.

\paragraph{Compute convention.}
We follow the FVCore convention used by DrivoR~\citep{drivor}, counting one
multiply--accumulate as one operation. Fused scaled dot-product attention
is excluded from this count. World4Scorer uses \gmacsOurs{} billion operations:
\gmacsOursEnc{} for the encoder and \gmacsOursHead{} for generation and scoring.
Including attention raises the total to \gmacsOursTrue{} billion; attention
over \visualTokens{} encoder tokens from four views accounts for almost all of
the \gmacsOursAttention{} billion added operations.

\paragraph{Scale.}
World4Scorer has \paramsSubmittedM{}M parameters, compared with \paramsViTLM{}M and
\paramsViTHM{}M for the ViT-L and ViT-H systems. Visual encoding dominates runtime.
Generation and scoring
take \latencyOursHeadMs{} of \latencyOursMs{}~ms per pass and
\backboneShareHead{} of compute including attention.

\paragraph{Precision.}
The latency column uses fp32 with TF32 disabled. On the same card and input,
the same graph runs in \latencyOursTfMs{}~ms with TF32 and
\latencyOursHalfMs{}~ms with fp16 (encoder: \latencyOursHalfEncMs{}~ms).
The $\latencyHalfSpeedup\times$ speedup from fp32 to fp16 reflects precision alone.
All reported planning scores use fp32.

\begin{table}[h]
\centering
\small
\setlength{\tabcolsep}{4pt}
\begin{tabular}{@{}llccrrr@{}}
\toprule
\appendixcolhead{Method} & \appendixcolhead{Image size} & \appendixcolhead{Cams} &
\appendixcolhead{Encoder} & \appendixcolhead{Params} & \appendixcolhead{GFLOPs} & \appendixcolhead{Latency} \\
\midrule
RAP-Dino \citep{rap} & $448\times768$ & 4 & ViT-H & 888M & 4760 & 690 \\
GTRS-Dense \citep{gtrs} & $512\times2048$ & 1 & ViT-L & 321M & 1730 & 400 \\
ZTRS \citep{ztrs} & $512\times2048$ & 2 & V2-99 & 81M & 840 & 193 \\
GTRS-Aug \citep{gtrs} & $512\times2048$ & 1 & V2-99 & 171M & 439 & 243 \\
GTRS-Dense \citep{gtrs} & $512\times2048$ & 1 & V2-99 & 81M & 404 & 96 \\
DrivoR \citep{drivor} & $672\times1148$ & 4 & ViT-S & 41M & 351 & 110 \\
\midrule
\textbf{Ours} & $672\times1148$ & 4 & ViT-S & \textbf{\paramsSubmittedM{}M}
  & \textbf{\gmacsOurs{}} & \textbf{\latencyOursMs{}} \\
\quad attention counted & & & & & \gmacsOursTrue{} & \\
\bottomrule
\end{tabular}
\caption{\textbf{Compute and parameter scale.} GFLOPs use the FVCore
convention of one operation per multiply--accumulate and exclude fused attention;
the \emph{attention counted} row includes it. Latency is in milliseconds
for one fp32 forward pass at batch one on an A100.
World4Scorer's latency excludes inertial re-ranking
(\latencyStateTermMs{}~ms on CPU for all 64 candidates).}
\label{tab:compute-scale}
\end{table}

\needspace{6\baselineskip}
\section{Evaluation Protocols}
\label{sec:appendix-evaluation}
We describe simulator labels and compare NAVSIM-v1 and NAVSIM-v2 objectives
with the model and candidate pools fixed, isolating the scoring rule.
\subsection{Simulator-Based Training Labels}
\label{sec:appendix-score-computation}

We obtain candidate labels by tracking each trajectory in the simulator and
measuring collision avoidance, progress, and other outcomes in three steps:
\begin{enumerate}
\setlength{\itemsep}{3pt}\setlength{\parskip}{0pt}
\item \textbf{Prepare the scene.} A model-independent cache stores the initial
ego state, centerline, on-route lane IDs, drivable area, and the PDM planner's
progress. It also stores $\pdmObsSteps$ occupancy timesteps as STRtree-indexed
polygons. Training and evaluation share this cache.
\item \textbf{Track the candidate.} Generated trajectories are detached before scoring, transformed to the global frame, and resampled to
$\pdmSimPoses$ poses at $\pdmSimInterval$\,s. An LQR controller tracks each
trajectory through a kinematic bicycle model, producing a
$\pdmSimHorizon$-second ego-state sequence.
\item \textbf{Measure the outcomes.} Collision avoidance, drivable-area
compliance, and driving direction use polygon queries against the cached maps.
The tracked ego-state sequence supplies time-to-collision, centerline progress, and comfort labels.
\end{enumerate}

In the simulator's aggregate score, the components combine in two groups. Collision avoidance and drivable-area
compliance multiply together, and their product both scales raw progress and
multiplies the aggregate score. For the v1 training-label scorer, progress,
time-to-collision, and comfort enter a weighted average whose weights match the
deployed NAVSIM-v1 tuple $\selWeightsVOne{}$; driving-direction compliance is
computed but has zero v1 weight.

\paragraph{Benchmark metrics.}
NAVSIM-v1 reports the PDM score~\citep{navsim},
\begin{equation*}
\mathrm{PDMS}=\mathrm{NC}\cdot\mathrm{DAC}\cdot
\frac{5\,\mathrm{TTC}+5\,\mathrm{EP}+2\,\mathrm{C}}{12}.
\end{equation*}
NAVSIM-v2 reports the extended PDM score~\citep{navsimv2},
\begin{equation*}
\mathrm{EPDMS}=\mathrm{NC}\cdot\mathrm{DAC}\cdot\mathrm{DDC}\cdot\mathrm{TLC}\cdot
\frac{5\,\mathrm{TTC}+5\,\mathrm{EP}+2\,\mathrm{HC}+2\,\mathrm{LK}+2\,\mathrm{EC}}{16},
\end{equation*}
which adds DDC and traffic-light compliance (TLC) as multipliers and lane keeping
(LK) and extended comfort (EC) to the weighted average; history comfort (HC)
replaces comfort (C). EC compares plans of consecutive frames. A term is
not penalized where the logged human trajectory also violates it.
Both metrics use the navtest scenes with non-reactive background agents.

\paragraph{Progress normalization.}
Progress is normalized by a reference distance:
\begin{itemize}
\setlength{\itemsep}{2pt}\setlength{\parskip}{0pt}
\item \textbf{Training:} the maximum of the candidate's raw progress and the cached planner progress.
\item \textbf{Evaluation:} the official scorer normalizes progress over the proposals passed
to it, which are the PDM reference and the selected trajectory in our evaluation.
\end{itemize}
A training label depends only on its trajectory and scene, not on other candidates.

\paragraph{Selection and runtime.}
At selection time, the predicted components are combined using
Equation~\ref{eq:ranking-utility}. Collision avoidance, drivable-area compliance,
and driving direction contribute weighted log probabilities; the remaining
components contribute the log of their weighted sum.
The NAVSIM-v1 weights $\selWeightsVOne{}$ assign zero weight to driving direction.

Generated candidates change at every step, requiring online labels for all
$N$ candidates per sample. Computing these labels on CPU with NumPy and
Shapely dominates training time.

\subsection{Differences Between NAVSIM Evaluation Objectives}
\label{sec:appendix-disagree}

NAVSIM-v1 and NAVSIM-v2 select different candidates in
\vOneVTwoArgmaxDisagree{} of scenes when applied to the same frozen
candidate pools of our planner. We use simulator sub-scores to isolate the evaluation
rule from prediction error. The two utilities have within-scene rank
correlation \vOneVTwoRho{}. Removing extended comfort raises
the correlation to \vOneVTwoRhoNoEc{} and reduces argmax disagreement to
\vOneVTwoArgmaxDisagreeNoEc{}. The V1-optimal candidate fails the EC check in \vOneVTwoEcFailShare{} of disagreement scenes. Selecting the V2-optimal candidate lowers native NAVSIM-v1 PDMS by
\vOneVTwoVOneCost{} points.

An exact additive decomposition attributes \vOneVTwoEcShare{} of the utility gap between the V1 and V2 choices to EC, \vOneVTwoDdcShare{} to driving direction (weighted only in V2), and \vOneVTwoRestShare{} to progress, time-to-collision, and comfort together
(progress alone: \vOneVTwoEpShare{}). The V2-optimal candidate gains \vOneVTwoEcRawDelta{} EC
points and loses \vOneVTwoEpRawDelta{} progress points relative to the V1-optimal candidate.

Progress equals $1.0$ for \vOneVTwoEpSaturated{} of candidates, so pool-wide
percentiles from the sixtieth upward coincide. In \jointInfeasible{} of scenes (\jointInfeasibleQFour{} in the most dynamic
quarter), no candidate is both top-quartile in progress and EC-passing.
The most dynamic quarter holds the scenes with the largest change between the
current and $+2$\,s front-view DINOv2 features. This share becomes
\jointInfeasibleMedian{} at the median threshold and \jointInfeasibleWithinScene{}
at a within-scene quartile. The within-scene progress--EC correlation is
$\epEcCorr{}$.

The recomputed V1 selections differ from the benchmark results only on
\vOneVTwoTieMismatch{} fp32 ties, changing mean PDMS by $6\times10^{-7}$.
Recomputed EC matches the benchmark on all \ecPairCount{} adjacent pairs.
The benchmark omits this term for the remaining \ecNoPairFrames{} frames, which lack a valid adjacent pair.

\needspace{6\baselineskip}
\section{Predictive Representations}
\label{sec:appendix-representations}
We examine visual targets and predictions, illustrate the content retained
by pooled features, and connect learned candidate states to score differences.
\subsection{Geometry of the Latent Prediction Target}
\label{sec:appendix-latent-geometry}

We compare current and future DINOv2 features over \latentBankSize{}
training frames. The current frame and its $+2.0$\,s
future are highly similar (mean cosine \latentPairedCosMean{}, fifth percentile
\latentPairedCosPFive{}), indicating that persistent scene content dominates the target and that a
measurable temporal change remains. Both feature sets have
effective rank \latentEffRank{} of \bankDim{} dimensions.

Pairing each current frame with a future from another scene reduces mean
cosine to \latentDerangedCosMean{}, close to the random-pairing baseline of \latentFloorCosMean{}; this shuffle
removes the pairing but keeps the target features. True and shuffled pairs are
readily distinguished (AUC \latentAUC{}, Cohen's $d$ \latentCohenD{}). Same-log
pairs are also less similar (\latentSameLogCos{}), so the true-pair advantage
is scene-specific.

\Needspace{8\baselineskip}
\subsection{Visualizing Predicted Features}
\label{sec:appendix-latent-decoding}

The visual readout predicts a \bankDim{}-dimensional front-view feature
$2.0$\,s ahead, mean-pooled from frozen DINOv2~\citep{dinov2} ViT-S/14 with registers~\citep{registers}. The logged trajectory supplies the only observed
future target; alternative queries share the predictor.

To visualize these features, we train two image reconstruction networks,
CondGrid and conditional latent diffusion, with the driving model frozen (Figure~\ref{fig:jepa-decode-pipeline}). Each network
combines the predicted feature with the current front view's
$16\!\times\!16$ DINOv2 patch grid, which supplies spatial detail.
This reconstruction stage is separate from the visual readout $g$ and is used
only for visualization. The alternative-query images below are qualitative
readouts; no observed future exists for these queries.

\Needspace{24\baselineskip}
\noindent\begin{minipage}[t]{0.52\textwidth}
\vspace{0pt}
The RAE decoder~\citep{rae} renders images from patch grids. Write $X_t$ for the current
patch grid and $\widehat u_{t,i}=g(z_{t,i})$ for a candidate's predicted visual
feature. CondGrid, a small conditional transformer with parameters $\eta$,
predicts a residual grid with classifier-free guidance at scale~2:
\[
\begin{aligned}
\Delta_{\mathrm{cfg}}
  &=\Delta_{\varnothing}(X_t)
    +2\!\left[\Delta_\eta(X_t,\widehat u_{t,i})
               -\Delta_{\varnothing}(X_t)\right],\\
\widehat X_{t+2\mathrm{s},i}&=X_t+\Delta_{\mathrm{cfg}},
\end{aligned}
\]
where $\Delta_{\varnothing}$ omits the visual feature. The frozen RAE ViT-XL
decoder renders $\widehat X_{t+2\mathrm{s},i}$.
Conditional latent diffusion uses cosine-schedule DDPM
$v$-prediction on noisy future grids. Inference uses
20-step DDIM (guidance 1, seed 17) and the RAE decoder.

Both reconstruction networks use the same \decodeTrainPairs{} navtrain current/future pairs.
Inputs are the observed current grid and future pooled vector; the target
is the full future grid. At test time, the pooled vector is replaced
by $\widehat u_{t,i}$, normalized to the global norm of the cached DINOv2 targets.
The driving model, DINOv2, and the RAE decoder stay fixed.
\end{minipage}\hfill
\begin{minipage}[t]{0.46\textwidth}
\vspace{0pt}
\centering
\includegraphics[width=\linewidth]{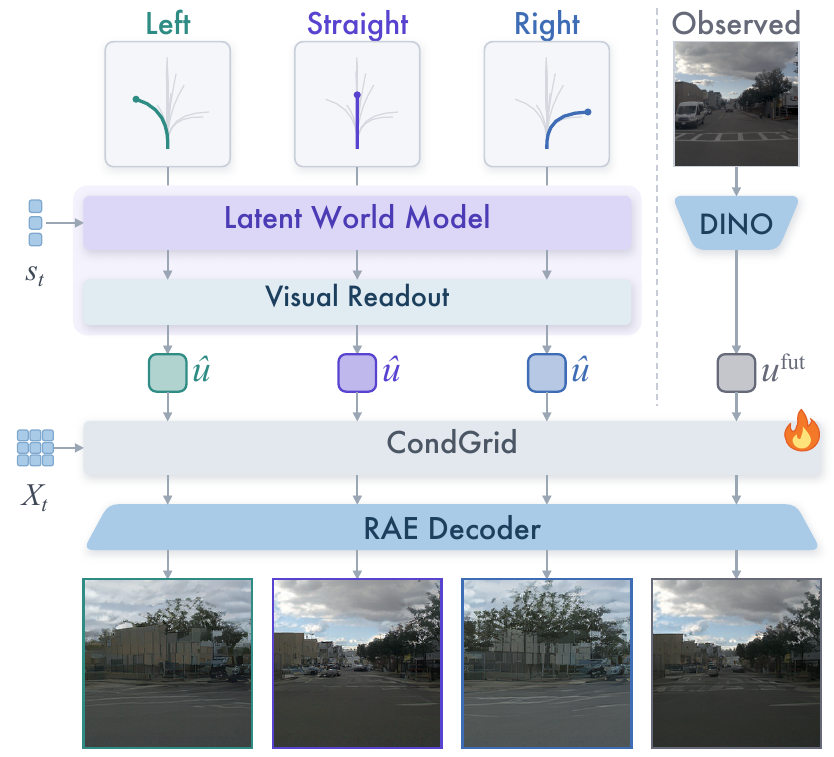}
\captionsetup{font=small,skip=3pt,justification=RaggedRight,singlelinecheck=false,width=\linewidth}
\captionof{figure}{\textbf{From query to image.} Scene, ego state and candidates are
fixed as the query changes. Each state $z$ yields a visual
feature $g(z)$, which the reconstruction network renders through a frozen decoder.
Left, Straight (selected), and Right match Figure~\ref{fig:latent-query-scan}.
The right column decodes the observed future's pooled feature.}
\label{fig:jepa-decode-pipeline}
\end{minipage}
\par\medskip

CondGrid uses token-cosine, MSE, and weighted change-direction losses;
diffusion uses a velocity loss. Both use scale perturbations, isotropic
direction noise, and 10\% vector dropout.

\paragraph{Changing the queried trajectory.}
Figure~\ref{fig:latent-query-scan} fixes six junctions and compares left-turn,
near-straight (model-selected), and right-turn candidates from each
64-candidate pool. The trajectory colors match the three decoded-image columns.
The three candidates' predicted features and the current DINOv2 grid pass through the same
CondGrid network and frozen RAE decoder. Changing the trajectory shifts the
decoded viewpoint while preserving scene structure. These views show responses to alternative plans; without matching
observations, their accuracy cannot be measured.

Removing the predicted vector preserves much of the image structure because
the reconstruction network still receives the current grid. At guidance 2, the observed
future's pooled vector improves regional patch cosine by \figfourNullGainDrivable{} to \figfourNullGainFar{} over this vector-free branch.
For the model-selected query, whole-grid cosine is \decodeGridCosGTwo{} at guidance 2 and
\decodeGridCosGOne{} at guidance 1, versus \decodeGridCosNull{} without the vector. The reconstruction therefore depends on both the pooled vector and the current
spatial grid; at guidance 2 the predicted vector does not raise whole-grid
similarity to the observed future.

\noindent
\begin{minipage}[t]{0.53\textwidth}
\vspace{0pt}
Current-view semantic masks partition the grids into road, dynamic objects,
and other static regions. Table~\ref{tab:regional-readout} pools patch-wise
cosine over the three query pairs, the patches in each region, and six scenes. The resulting similarities measure how much the reconstructed
regions respond to different queries in feature space, before the decoder renders an image.
\end{minipage}\hfill
\begin{minipage}[t]{0.43\textwidth}
\vspace{0pt}
\centering
\small
\begin{tabular*}{\linewidth}{@{\extracolsep{\fill}}lr@{}}
\toprule
\appendixcolhead{Region} & \appendixcolhead{Grid cosine} \\
\midrule
Road & \queryRoadConsistency{} \\
Dynamic objects & \queryDynamicConsistency{} \\
Other static & \queryStaticConsistency{} \\
\bottomrule
\end{tabular*}
\captionsetup{font=small,skip=3pt,justification=RaggedRight,singlelinecheck=false,width=\linewidth}
\captionof{table}{\textbf{Cross-query readout consistency.} Higher cosine
means less variation between query-conditioned grids.}
\label{tab:regional-readout}
\end{minipage}
\par\medskip

\begin{figure}[!htbp]
\centering
\includegraphics[width=\textwidth]{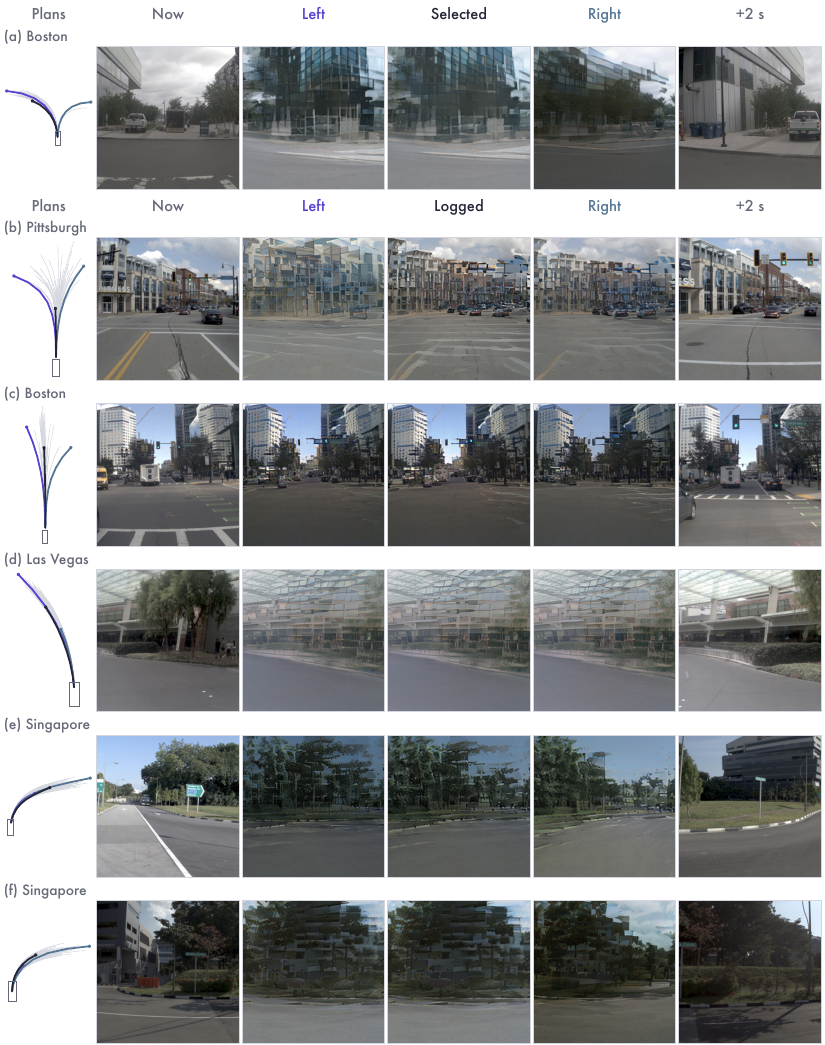}
\captionsetup{font=small,skip=3pt,justification=RaggedRight,singlelinecheck=false}
\caption{\textbf{Additional visual readouts under alternative plans.}
Each row shows the queried trajectories, current photograph, three visual
readouts, and observed future ($+2$\,s). Left and Right denote relative
lateral alternatives, including variants of the same turn. The center query
is model-selected in (a) and logged in (b--f); observed futures always follow
the logged path in that scene. Alternative-plan images are qualitative readouts
from the separately trained reconstruction network.}
\label{fig:appendix-query-examples}
\end{figure}

\paragraph{Additional road scenes.}
Figure~\ref{fig:appendix-query-examples} adds six locations absent from
Figure~\ref{fig:latent-query-scan}, covering broad junctions, urban streets,
and curved roads. Changing the query shifts the road opening and visible
building surfaces, with subtler differences between nearby plans.
All eighteen images are rendered with the same fixed CondGrid network and RAE decoder.

\begin{figure}[!htbp]
\centering
\includegraphics[width=\textwidth]{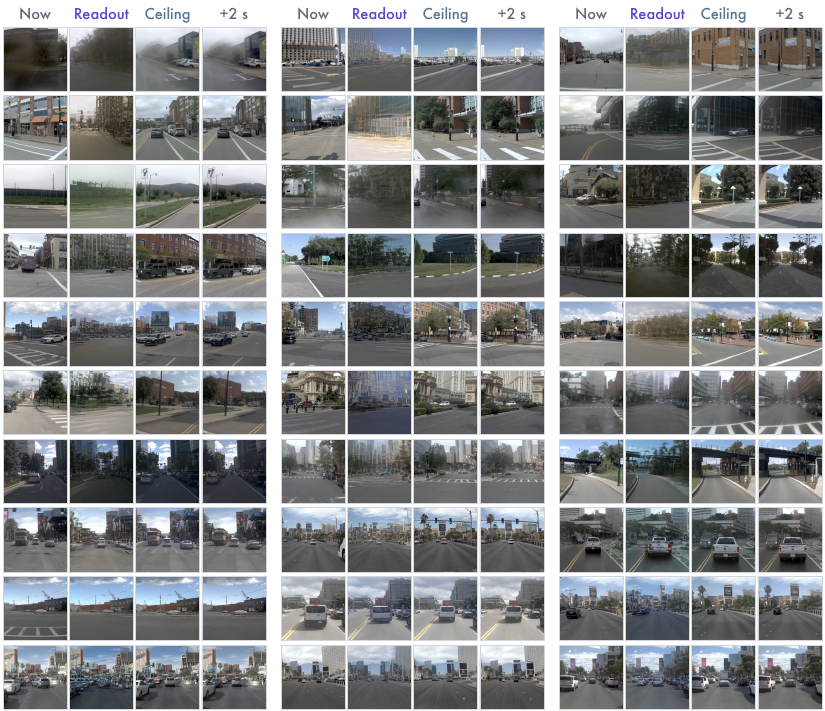}
\captionsetup{font=small,skip=3pt,justification=RaggedRight,singlelinecheck=false}
\caption{\textbf{Complete post-hoc decoding gallery.} Thirty pre-specified
navtest scenes, three per row, sorted by decreasing current-to-future DINOv2 feature change.
Each group shows the current frame (Now), executed-query readout, full-grid
decoder ceiling, and observed future ($+2$\,s).
Logged-query reconstructions combine predicted features with the current spatial
grid through the separately trained reconstruction network. The ordering is independent of
decoding quality.}
\label{fig:latent-decoding-full}
\end{figure}

\Needspace{8\baselineskip}
\paragraph{Reading the full gallery.}
Figure~\ref{fig:latent-decoding-full}
shows all 30 pre-specified navtest scenes, ordered by decreasing change between
current and future DINOv2 features. The columns show the current photograph,
the logged-query reconstruction of
the model's predicted feature, a decoder ceiling obtained by feeding the
observed future's full patch grid directly to the same frozen RAE decoder,
and the observed future. The ceiling isolates decoder quality; the reconstruction
also depends on how much spatial information survives pooling and is captured by future prediction.

Low-change scenes are legible because the current grid supplies their spatial
structure. With larger viewpoint changes, the pooled feature leaves several
possible spatial layouts unresolved.

CondGrid is deterministic; diffusion samples from the same fixed feature.
On the same 30 gallery scenes, diffusion yields sharper images but lower
feature and pixel fidelity. Sampling occurs in the reconstruction network; JEPA emits one
vector per query. The fixed-pool interventions in
Appendix~\ref{sec:appendix-pool-audit} assess how candidate states affect
trajectory ordering and selection performance.

\subsection{Candidate-State Geometry}
\label{sec:appendix-latent-content}

We examine the candidate states used by the scorer.
Across 128 navtrain scenes with 64 bank candidates each, the 256-dimensional
states $z$ have effective rank \sharedStateEffRank{}. Within-scene candidate
differences account for \sharedStateWithinShare{}\% of total state variance.

\begin{figure}[htbp]
\centering
\includegraphics[width=\linewidth]{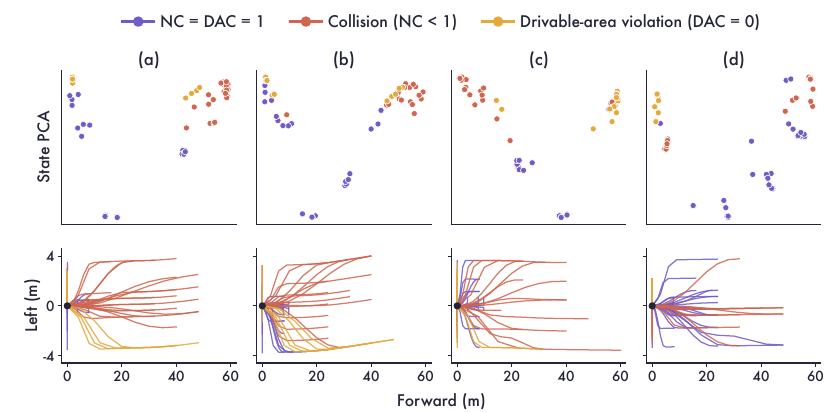}
\captionsetup{font=small,skip=4pt,justification=RaggedRight,singlelinecheck=false}
\caption{\textbf{Candidate states within four scenes.}
Top: the first two principal components of the 64 candidate states in each
scene, which explain \sharedStateShownPcaLow{}--\sharedStateShownPcaHigh{}\%
of their variance. Bottom: the same candidates' bank trajectories in the ego
frame, with axes not to scale. Candidates with both a collision and a
drivable-area violation are colored as drivable-area violations. The four
scenes come from different logs and were chosen for balanced outcome groups
(at least \sharedStateShownMinGroup{} candidates each). Vertical segments at
the origin are bank trajectories that pair zero forward speed with a lateral
transition.}
\label{fig:scene-states}
\end{figure}

The scorer compares candidates within a scene, so we examine the states of
each scene separately. The 128 scenes come from 16 navtrain logs, eight per log. We group candidates by simulator outcome: no collision and no
drivable-area violation (NC $=$ DAC $=1$), collision (NC $<1$), or
drivable-area violation (DAC $=0$). On average, \sharedStateKnnAgree{}\% of a
candidate's five nearest states in its scene share its outcome group, compared
with \sharedStateKnnTraj{}\% for the five nearest trajectories and
\sharedStateKnnShuffled{}\% after shuffling outcome labels within each scene.
The state agreement exceeds the shuffled value in all 128 scenes and the
trajectory value in \sharedStateAboveTraj{} of them. Figure~\ref{fig:scene-states}
shows four scenes.

\subsection{Supervision Geometry and Candidate Comparisons}
\label{sec:appendix-anchor-contrast}

We expand the local model of Section~\ref{sec:method-predictive} and prove
Proposition~\ref{prop:anchor-contrast}. The analysis follows the principle
that supervision should constrain differences relevant to candidate selection,
as in transductive experimental design~\citep{fiez2019transductive}.
Here the observation maps are the network's visual and outcome readouts,
whose parameters are themselves learned.

\paragraph{Observation maps and candidate context.}
Fix scene tokens, encoded ego states, trajectory coordinates, and the candidate
set used by each predictor call. Let $\theta\in\mathbb R^p$ collect the
parameters of $\phi,P,g,H$. For an executed query, define the normalized
visual prediction
\[
v_n(\theta)=\frac{g_\theta(z_n^{\rm exec}(\theta))}
{\|g_\theta(z_n^{\rm exec}(\theta))\|_2}.
\]
We work around a fitted $\theta_0$ at which these predictions are nonzero and
the readouts are differentiable; we do not assume a stationary training loss. Stack the
executed-query predictions into $F(\theta)$ and the supervised outcome logits
into $O(\theta)$, with Jacobians $J_{\rm fut}$ and $J_{\rm out}$ at $\theta_0$. Normalization follows the cosine loss (Equation~\ref{eq:factual-jepa});
only the latent state's readout $g(z)$ is compared with the observed future.

Each state retains the context of its actual predictor call: an executed
query is evaluated alone, generated queries jointly, and bank queries in a
separate set. Shared weights couple these calls. For a candidate pair $(i,j)$ in a fixed scene and pool, define
$\hat\Delta_{ij}(\theta)=\hat J_i(\theta)-\hat J_j(\theta)$ using
Equation~\ref{eq:ranking-utility}, and
$c_{ij}=\nabla_\theta\hat\Delta_{ij}(\theta_0)$.
This pool may come from a new scene or contain unsupervised queries.

\paragraph{Training-loss geometry.}
The first-order changes under $\theta_0+\delta$ are $J_{\rm fut}\delta$,
$J_{\rm out}\delta$, and $c_{ij}^\top\delta$. The generalized Gauss--Newton (GGN)
surrogate of the outcome and future terms of Equation~\ref{eq:training-objective}
has the curvature term $\tfrac12\delta^\top(J_{\rm out}^\top W_{\rm out}J_{\rm out}
+\lambda_{\rm fut}J_{\rm fut}^\top W_{\rm fut}J_{\rm fut})\delta$.
Here $W_{\rm out}$ holds the logit-space curvature $w_i\,p(1-p)$ of each binary
cross-entropy term, including soft-label progress. The bank's progress term is
an L1 loss on $p_{\rm EP}=\sigma(\ell_{\rm EP})$; it is piecewise linear in
$p_{\rm EP}$ but not in the logit, and we omit it from the logit-space GGN, so
$G_{\rm out}$ is a partial curvature surrogate rather than the exact curvature of
the full mixed loss. On unit vectors the cosine
loss equals $\tfrac12\|v_n-\bar u_n\|_2^2$ with
$\bar u_n=u_n^{\rm fut}/\|u_n^{\rm fut}\|_2$, so $W_{\rm fut}=I/M$.
Write $A_O=W_{\rm out}^{1/2}J_{\rm out}$, $A_F=W_{\rm fut}^{1/2}J_{\rm fut}$, and
$\lambda=\lambda_{\rm fut}$. For $\mu,\epsilon,\lambda>0$, define
\begin{equation}
\mathcal E_\epsilon(G)=\left\{\delta:
\mu\|\delta\|_2^2+\|A_O\delta\|_2^2
+\lambda\|A_F\delta\|_2^2\le\epsilon^2\right\}.
\label{eq:local-output-set}
\end{equation}
The quadratic form is the positive-semidefinite curvature part of this local
surrogate plus a trust-region term $\mu\|\delta\|_2^2$ that keeps $\delta$ where
the linearization holds. It defines the trust region of a first-order
sensitivity analysis; it is not the complete change in the mixed training loss,
whose first-order and residual-curvature terms are omitted. The budget
$\epsilon$ bounds the change; neither $\mu$ nor $\epsilon$ enters
training or inference. Equivalently,
$\mathcal E_\epsilon(G)=\{\delta:\delta^\top G\delta\le\epsilon^2\}$ with the
positive-definite matrix
\begin{equation}
G=G_O+\lambda A_F^\top A_F,\qquad G_O=\mu I+A_O^\top A_O.
\label{eq:local-supervision-geometry}
\end{equation}
All comparisons below hold at the same $\theta_0$ and perturbation budget.
Define the maximal first-order change in a candidate score difference, or radius, as
\begin{equation}
\rho_{ij}(G):=\sup_{\delta\in\mathcal E_\epsilon(G)}|c_{ij}^\top\delta|
=\epsilon\sqrt{c_{ij}^\top G^{-1}c_{ij}}.
\label{eq:anchor-contrast-radius}
\end{equation}
The main-text quantities are $\rho_{ij}^{\rm out}=\rho_{ij}(G_O)$ and
$\rho_{ij}^{\rm out+fut}=\rho_{ij}(G)$. Thus both use the same model and
budget; only the added visual-output constraints differ.

\paragraph{Proof of Proposition~\ref{prop:anchor-contrast}.}
Set $v=G^{1/2}\delta$, so $\|v\|_2\le\epsilon$. Cauchy--Schwarz yields
\[
|c_{ij}^\top\delta|
=|(G^{-1/2}c_{ij})^\top v|
\le\epsilon\sqrt{c_{ij}^\top G^{-1}c_{ij}}.
\]
For $c_{ij}\ne0$, equality is attained at both signs of
\[
\delta^*=\frac{\epsilon G^{-1}c_{ij}}
{\sqrt{c_{ij}^\top G^{-1}c_{ij}}}.
\]
Scaling attains every intermediate value; the case $c_{ij}=0$ has radius zero.
This proves Equation~\ref{eq:anchor-contrast-radius}.

To establish strict contraction, set $b=A_FG_O^{-1}c_{ij}$.
The Woodbury identity gives
\begin{equation}
\rho_{ij}(G_O)^2-\rho_{ij}(G)^2
=\epsilon^2 b^\top
\left(\lambda^{-1}I+A_FG_O^{-1}A_F^\top\right)^{-1}b.
\label{eq:visual-contrast-contraction}
\end{equation}
The middle matrix is positive definite. Hence the reduction is nonnegative,
proving Equation~\ref{eq:complementary-supervision},
and is strictly positive exactly when $A_FG_O^{-1}c_{ij}\ne0$.
For nonzero $c_{ij}$, the maximal-change direction under outcome constraints
is proportional to $G_O^{-1}c_{ij}$, so this is precisely the condition that
visual predictions vary along that direction. Because $W_{\rm fut}$ is
invertible, it is the condition $J_{\rm fut}G_{\rm out}^{-1}c_{ij}\ne0$ of
Proposition~\ref{prop:anchor-contrast}, with $G_{\rm out}=G_O$ and
$G_{\rm fut}=J_{\rm fut}^\top W_{\rm fut}J_{\rm fut}$.
This completes the proof. The inverse $G_O^{-1}$ emphasizes comparison
directions weakly constrained by the existing outcome observations; $A_F$
determines whether those same directions are also constrained by visual supervision.

\paragraph{Score differences and selection.}
The linearized score difference spans the interval
\[
[\hat\Delta_{ij}(\theta_0)-\rho_{ij}(G),\,
  \hat\Delta_{ij}(\theta_0)+\rho_{ij}(G)].
\]
Every local model orders $i$ strictly above $j$ if and only if
$\hat\Delta_{ij}(\theta_0)>\rho_{ij}(G)$.
Consequently, a fitted winner $k$ remains the unique winner throughout
$\mathcal E_\epsilon(G)$ if and only if this inequality holds for every
$j\ne k$. The condition allows nonzero variation in both score values and
predictive states, provided it does not reverse a winning comparison between candidates.

With exact first-order output preservation, the unconstrained directions
are $\mathcal N=\ker A_F\cap\ker A_O$.
A comparison is fixed along every direction in $\mathcal N$ precisely when
$c_{ij}\perp\mathcal N$, or equivalently when $c_{ij}$ belongs to the
row space of the vertically stacked observation Jacobians.
This is the zero-output-change counterpart of the finite-radius analysis of candidate comparisons.

\paragraph{Additional candidate outcomes.}
Let $B$ stack the Jacobians of outcome predictions at newly supervised
candidate queries, with their fixed weights absorbed into its rows.
Adding these constraints yields $G_+=G+B^\top B$. Applying the same
identity gives
\begin{equation}
\begin{split}
\rho_{ij}(G)^2-\rho_{ij}(G_+)^2
&=\epsilon^2 a^\top(I+BG^{-1}B^\top)^{-1}a,\\
a&=BG^{-1}c_{ij}.
\end{split}
\label{eq:outcome-anchor-contraction}
\end{equation}
The reduction is positive exactly when $BG^{-1}c_{ij}\ne0$. Additional outcome
labels therefore help when they constrain comparison directions that existing
supervision leaves unresolved.

\paragraph{Relation to learning.}
The result characterizes local sensitivity at a fitted network under
the same quadratic perturbation budget, not the optimization trajectory or
global planning performance after training.
Its quadratic form is a partial generalized Gauss--Newton metric for the BCE
and cosine terms, not the complete mixed training objective. No stationarity
of the individual loss terms is assumed; the analysis concerns local
first-order score sensitivity under this metric, and $\mu$ restricts it to the
region where the linearization holds.
Target values affect the fitted network and hence its local geometry;
replacing a target at fixed parameters does not itself change the output
Jacobian. Section~\ref{sec:rq-components} evaluates the learning effects
of future targets and correctly paired candidate outcomes. The analysis
identifies a condition for complementary supervision without assuming
that latent-state coordinates or complete alternative futures are recovered.

\needspace{6\baselineskip}
\section{Candidate States and Trajectory Selection}
\label{sec:appendix-selection}
With fixed trained weights, we measure how state reassignment affects selection
and describe the OGBench-Cube setup with the LeWM planner.
\subsection{Candidate-State Interventions with Fixed Pools}
\label{sec:appendix-pool-audit}

With the checkpoint and pool fixed, Table~\ref{tab:full-functional-audit}
compares matched states, scene-mean replacement, and within-scene reassignment
on the full test set. The common oracle is 99.33 PDMS; selected PDMS equals this oracle minus the
regret.

\begin{table}[!htbp]
\centering
\begin{minipage}{\columnwidth}
\centering
\small
\setlength{\tabcolsep}{6pt}
\renewcommand{\arraystretch}{0.98}
\begin{tabular*}{\linewidth}{@{\extracolsep{\fill}}lrr@{}}
\toprule
\appendixcolhead{Candidate-state intervention} & \appendixcolhead{Oracle regret (95\% CI)} &
\appendixcolhead{$\Delta$ vs. matched (95\% CI)} \\
\midrule
Matched candidate state & \rSixMatchedRegret{} \rSixMatchedRegretCI{} & -- \\
Scene-mean replacement & \rSixSceneMeanRegret{} \rSixSceneMeanRegretCI{} &
\rSixSceneMeanDelta{} \rSixSceneMeanCI{} \\
Within-scene derangement & \rSixDerangedRegret{} \rSixDerangedRegretCI{} &
\rSixDerangedDelta{} \rSixDerangedCI{} \\
\bottomrule
\end{tabular*}
\captionsetup{width=\linewidth,justification=RaggedRight,singlelinecheck=false}
\caption{\textbf{Candidate-state interventions.} Every
intervention uses World4Scorer and the same
\figTwoNumCandidates{} candidates of the released DrivoR model for all \navtestSize{} NAVSIM-v1 test scenes,
so matched PDMS is not the generator-pool result of Table~\ref{tab:navsim-v1-context}.
Regret is measured relative to the common pool oracle; positive differences are
worse. Paired intervals bootstrap over the \figTwoNumLogs{} evaluation logs.}
\label{tab:full-functional-audit}
\end{minipage}
\end{table}

Replacing all candidate states with their scene mean gives every candidate
the same score, so the first-index tie rule selects one fixed pool index on
all \navtestSize{} scenes. This row measures selection without a ranking
signal. Reassignment preserves states but changes their assigned candidates.

\appendixfinding{State reassignment disrupts trajectory selection.}{%
Reassignment changes the selected candidate on \rSixDerangedFlipRate{} of
scenes and reduces ordering accuracy to \rSixDerangedOrderAcc{}, approximately
chance. Together with the higher regret, this shows that selection depends
on the candidate--state correspondence.}

\paragraph{Variation across reassignments.}
All 100 random reassignments reduce selected PDMS. Their means range from
\rSixRepeatMin{} to \rSixRepeatMax{}, averaging \rSixRepeatSelected{}
(95\% log-cluster interval \rSixRepeatSelectedCI{}). The average loss against
matched states is \rSixRepeatDrop{} points (paired interval \rSixRepeatDropCI{}).

These repeats use the epoch-23 checkpoint and all \navtestSize{} scenes.
Each draws one fixed-point-free permutation of the 64 indices and applies it
across scenes, using seeds 20260909--20261008. The score heads read each state plus the same encoded ego state, so
reassigning states permutes the candidate scores.
Every repeat uses the same native-V1 utilities and first-index tie rule.

We average repeats within each scene, then form intervals from 100,000
bootstrap draws of the \figTwoNumLogs{} logs. Because all repeats share one checkpoint, this variation isolates the
reassignment draw.

\paragraph{Comparison with DrivoR.}
Reassigning DrivoR's scorer-attention outputs raises regret by \xlinDerangedDelta{} points over its matched \xlinRegretMatched{}
(95\% log-cluster interval \xlinDerangedCI{}) and reduces ordering accuracy
from \xlinOrderMatched{} to \xlinOrderDeranged{}. State assignment thus
matters even without future prediction.

The control matches the World4Scorer pools and native-V1 utilities across
\navtestSize{} scenes and \figTwoNumLogs{} logs, with permutation seed
20260802, first-index ties, and 100,000 log-bootstrap draws.

\paragraph{Selected scenes.}
Figure~\ref{fig:representation} shows twelve scenes from eleven logs: four
left turns, four straight paths, and four right turns. The scenes show
diverse geometry, distinct paths, and agreement with the driver.
Matched states pick a pool-best path in every case. Purple stars mark these
selections; amber dotted paths show the driver. Paths and maps use real, equally scaled coordinates.

\subsection{Outcome-Based Scoring Inside the LeWM Planner}
\label{sec:appendix-cube}

We use the official LeWM~\citep{lewm} implementation on
OGBench-Cube~\citep{ogbench}. Its world model predicts the result of an action
sequence, and its planner searches for a sequence that reaches the goal.
We freeze the world model and add outcome-based scoring to this search.

The planner runs cross-entropy-method (CEM) search over \cubeCandidates{} action
sequences of \cubeHorizonBlocks{} blocks, each block \cubeFrameskip{}
environment steps, and keeps the \cubeElite{} best sequences at every
iteration. The goal is the expert's cube pose \cubeGoalOffset{} steps ahead,
and each episode has a budget of \cubeBudget{} environment steps. Success requires
placing the cube within \cubeToleranceCm{}\,cm of this goal pose.

LeWM scores each sequence by the distance between its predicted final feature
and the goal feature, favoring smaller distances. Our scorer takes these two
features, their distance, and the action sequence, and predicts its outcome
score. At each CEM iteration, we standardize the negative distance and the learned score across candidates
and add the score to the distance with weight
$\lambda_{\rm cube}=\cubeBlendC{}$. This combined score selects the retained
sequences under the same world model and search budget.

The scorer learns from simulator outcomes of candidates collected by the
LeWM planner, using training seeds separate from evaluation. Both planners
use \cubeItersAppD{} CEM iterations, following LeWM's Appendix~D.
We choose $\lambda_{\rm cube}$ by success on tuning seeds \cubeTuneSeeds{};
these seeds are separate from the reported evaluation. Table~\ref{tab:cube-main}
reports the mean over seeds 0--9 and \cubeSeedMain{}, LeWM's released
evaluation seed, each with \cubeEpisodes{} paired episodes. Each pair shares an initial state and goal.

\paragraph{Evaluation seeds.}
All seeds use the same trained scorer, weight, and \cubeItersAppD{}-iteration CEM budget.
LeWM succeeds on \cubeAnchorTenSucc{}/\cubeTenEpisodes{} episodes
(\cubeAnchorTenMean{}\%) and outcome-based scoring on
\cubeOursTenSucc{}/\cubeTenEpisodes{} (\cubeOursTenMean{}\%), a gain of
\cubeDeltaTen{} percentage points. The scorer rescues \cubeTenRescued{} episodes
and loses \cubeTenLost{}; success improves on \cubeSeedsUpTen{} seeds, ties on
\cubeSeedsTieTen{}, and decreases on \cubeSeedsDownTen{}. These runs vary the
evaluation initial states and goals while the world model and scorer stay fixed.

\paragraph{Scorer architecture and training.}
The scorer input concatenates the layer-normalized predicted final and goal features,
their difference and elementwise product, the log-transformed LeWM distance, and
the mean and standard deviation of the action sequence (819 dimensions).
A layer-normalized MLP with GELU activations and dropout 0.05 maps this input
through widths 256 and 128 to a scalar. The target is the negated
minimum cube-to-goal distance along each candidate's simulator rollout,
standardized within its candidate group. The loss adds smooth-$L_1$ regression and a
listwise cross-entropy term with weight 0.25 and softmax temperature 0.5.

Training candidates come from LeWM planning runs with seeds 100--107.
Episodes used in any evaluation are excluded, as are candidate groups whose
target standard deviation is below $10^{-4}$. An episode-disjoint split
gives 2,982 training groups from 262 episodes and 725 validation groups from
65 episodes. We train for 80 epochs with AdamW (learning rate $10^{-3}$,
weight decay $10^{-4}$, cosine schedule, 64 groups per batch) and keep the
epoch with the highest validation pairwise ordering accuracy (epoch 6,
counting from zero).

\needspace{6\baselineskip}
\section{Sequential Planning and Closed-Loop Evaluation}
\label{sec:appendix-sequential}
We analyze failures between consecutive selections, measure the effect of
inertial re-ranking and its history input, and describe the Bench2Drive
closed-loop evaluation.
\subsection{Extended-Comfort Failure Taxonomy}
\label{sec:appendix-ec-taxonomy}

Table~\ref{tab:ec-taxonomy} classifies our planner's \ecFailCount{} framewise
extended-comfort failures among \ecPairCount{} consecutive pairs, according
to the alternatives available at each decision:
\begin{itemize}
\setlength{\itemsep}{2pt}\setlength{\parskip}{0pt}
\item \textbf{Score--continuity conflict:} a passing alternative exists, but the selected
failing candidate has higher utility by more than the near-tie threshold.
\item \textbf{Near-tie:} a passing alternative is within
$\varepsilon=4.1\times10^{-5}$ of the selected utility.
\item \textbf{Pool-limited:} no candidate passes the EC check.
\end{itemize}
Score--continuity conflicts account for \ecTradeShare{} of failures,
whereas near-ties account for \ecTieShare{}. Most pool-limited failures (\ecPoolQFour{}) fall in the most dynamic quarter
of pairs (Appendix~\ref{sec:appendix-disagree}). After inertial
re-ranking, \ecResidualPool{} of residual failures have no compatible candidate in the pool.

For DrivoR, $7.3\%$ of pairs lack compatible candidates, so no selection
from its pools can exceed an EC of \ecCeilingReleased{}.

The EC check also rejects some logged driving: human plans
pass on \humanEcPass{}\% of all adjacent pairs and on
\humanEcInPool{}\% of our pool-limited pairs. Thus part of the remaining
failure rate comes from motion-profile differences that also occur in logged human driving.

\begin{table}[H]
\centering
\begin{minipage}{0.72\columnwidth}
\centering
\footnotesize
\setlength{\tabcolsep}{6pt}
\renewcommand{\arraystretch}{1.05}
\begin{tabular*}{\linewidth}{@{\extracolsep{\fill}}lrr@{}}
\toprule
\appendixcolhead{Failure class} & \appendixcolhead{Share of failures} & \appendixcolhead{Repaired by $\Psi$} \\
\midrule
Score--continuity conflict & \ecTradeShare{} & \ecTradeRepair{} \\
Near-tie & \ecTieShare{} & \ecTieRepair{} \\
Pool-limited & \ecPoolShare{} & \ecPoolRepair{} \\
\bottomrule
\end{tabular*}
\captionsetup{width=\linewidth,justification=RaggedRight,singlelinecheck=false}
\caption{\textbf{Consecutive-choice failures.} Classification of \ecFailCount{} framewise EC failures among \ecPairCount{} consecutive pairs.}
\label{tab:ec-taxonomy}
\end{minipage}
\end{table}

\subsection{Changes in Consecutive Selections}

With our planner's weights and candidates fixed, inertial re-ranking changes
\ecFlipPairs{} consecutive pairs from EC failure to pass and \ecRegressPairs{}
from pass to failure. On the repaired pairs, re-ranking reduces heading jitter by \ecFlipHeadingDrop{},
lateral discrepancy by \ecFlipLateralDrop{}, and the $L_2$ distance between
consecutive plans by \ecFlipPathDrop{}. Among changed selections, distance to
the logged plan drops from \stateGtBefore{} to \stateGtAfter{}\,m.

\subsection{Re-Ranking with Fixed Candidate Pools}
\label{sec:appendix-state-attr}

Inertial re-ranking also improves DrivoR: with DrivoR's weights and candidates
fixed, it raises EPDMS from \epdmsReleasedSameHost{} to \epdmsReleasedState{},
a \stateGainReleased{}-point gain from changing the final selection rule alone.

\paragraph{Sensitivity to the state weight.}
The reported results fix $\lambda_{\rm state}=1$. We re-run inertial re-ranking at $\lambda_{\rm state}\in\{0.25,0.5,1,2\}$ on
the fixed pool and scores of Coverage, a variant of our planner whose generator
is also trained to cover high-scoring bank trajectories. The four runs give \ecLamQuarter{},
\ecLamHalf{}, \ecLamOne{}, and \ecLamTwo{} EPDMS: varying the weight eight-fold moves EPDMS by at most \ecLamBand{} points. The gain over framewise selection
persists throughout this range of compatibility-penalty weights.

\subsection{History Corruption Controls}
\label{sec:appendix-history}

Using the correct previous plan improves EPDMS by \historyCorrectWrongGap{} points over
wrong-log history (95\% CI
[\historyCorrectWrongCILow{},~\historyCorrectWrongCIHigh{}]) and by
\historyCorrectStaleGap{} points over two-step-old history
(95\% CI [\historyCorrectStaleCILow{},~\historyCorrectStaleCIHigh{}]).
Table~\ref{tab:history-controls} uses the same fixed Coverage pool and scores
as Appendix~\ref{sec:appendix-state-attr} and changes only the history supplied
to the EC check, testing whether the benefit comes from the immediately
preceding plan. Its last row instead resubmits the previous frame's selected
plan without choosing among the current candidates; it scores below framewise
selection, so the gain does not come from repeating plans.

\begin{table}[!htbp]
\centering
\begin{minipage}{0.82\columnwidth}
\centering
\small
\setlength{\tabcolsep}{5pt}
\begin{tabular*}{\linewidth}{@{\extracolsep{\fill}}lr@{}}
\toprule
\appendixcolhead{Previous plan supplied} & \appendixcolhead{V2 EPDMS} \\
\midrule
Memoryless & \epdmsSupportExp \\
Wrong-log previous plan & \epdmsSupportExpWrongHistory \\
Same-log, two-step-old plan & \epdmsSupportExpStaleHistory \\
Correct previous plan & \epdmsSupportExpEC \\
Copy previous plan & \epdmsCopyPrev \\
\bottomrule
\end{tabular*}
\captionsetup{width=\linewidth,justification=RaggedRight,singlelinecheck=false}
\caption{\textbf{History-source controls for inertial re-ranking} on
Coverage. Memoryless is framewise selection; Copy previous plan resubmits the previous frame's plan. Differences use unrounded EPDMS.}
\label{tab:history-controls}
\end{minipage}
\end{table}

\Needspace{8\baselineskip}
\subsection{Closed-Loop Evaluation on Bench2Drive}
\label{sec:b2d-closed-loop}

\begin{figure}[!htbp]
\centering
\includegraphics[width=0.97\textwidth]{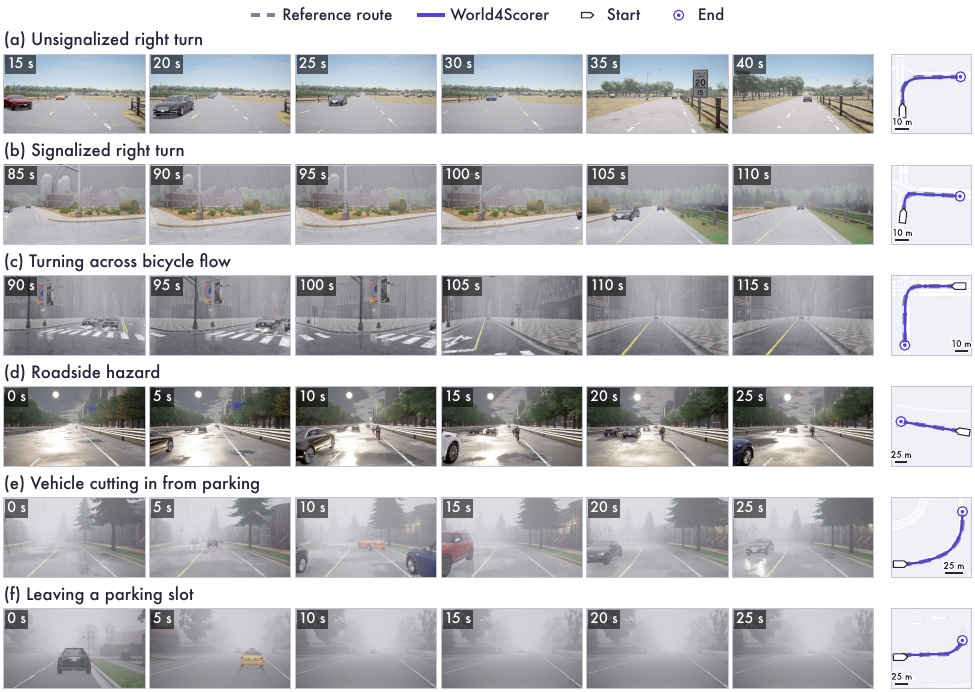}
\captionsetup{width=\textwidth,justification=RaggedRight,singlelinecheck=false}
\caption{\textbf{World4Scorer in six Bench2Drive scenarios.}
Each row shows six consecutive camera views at 5\,s intervals around a driving
maneuver. The small map at right shows the full reference route (dashed) and
driven trajectory (purple). All six routes are completed by the Ours
configuration of Table~\ref{tab:b2d-ds}, with Driving Score and Route Completion
both equal to 100.}
\label{fig:b2d-showcase}
\end{figure}

We train the planner on the official 1000-clip subset and evaluate it in
CARLA using Bench2Drive~\citep{bench2drive}. We evaluate an adapted system; published baselines are not retrained under
matched conditions. We initialize from a route-blind checkpoint trained on the
same clips, add the route-point input, and
use the simulator and route inputs below. Table~\ref{tab:b2d-ds}
compares Driving Scores, and Figure~\ref{fig:b2d-showcase} shows six completed
scenarios with camera frames every five seconds.

The CARLA ego input adds the route point \bTwoDRouteTargetDist{}\,m ahead,
expressed in ego-frame meters, to the NAVSIM pose, velocity, acceleration,
and command inputs. The same coordinate conversion supplies this route
point during training and deployment.

We choose the distance by scanning 10--40\,m over \bTwoDRouteTargetScanFrames{}
frames from \bTwoDRouteScanClips{} training clips. At \bTwoDRouteTargetDist{}\,m, the
point lies on the side of the turn for \bTwoDTurnSignLeft{} of left-turn and
\bTwoDTurnSignRight{} of right-turn frames, while the route ends before the point on
only \bTwoDRouteEndShare{} of frames. We then add two zero-initialized input
columns to the route-blind checkpoint and retain its training data, loss, and
schedule.

\paragraph{Training.} The model and losses follow NAVSIM with two changes.
Outcome labels are computed against the logged agent boxes and a drivable-area
raster of each clip, and extra negatives come from a trajectory vocabulary
instead of the CLOVER bank. The future loss has weight \bTwoDFutWeight{}, and
its gradient into the predictor is scaled by \bTwoDFutGradScale{}. Both the
route-blind and the route-point model train for \bTwoDEpochs{} epochs with AdamW
at a learning rate of \bTwoDLR{} and \bTwoDBatch{} clips per GPU.

\Needspace{24\baselineskip}
\noindent\begin{minipage}[t]{0.52\textwidth}
\vspace{0pt}
Two deployment rules need no retraining:
\begin{itemize}
\setlength{\itemsep}{2pt}\setlength{\parskip}{0pt}
\item \textbf{Command retention} keeps a route node active until the vehicle
passes it, preserving the turn command through an intersection.
\item \textbf{Route re-ranking} measures agreement with the route polyline
within 30\,m using an exponential kernel of average waypoint-to-route distance,
weighted toward later waypoints. We standardize this agreement across
candidates and add it to the learned utility with weight
$4\sigma$, where $\sigma$ is the pool's utility standard deviation.
\end{itemize}
We select the plan with the highest combined score.
\end{minipage}\hfill
\begin{minipage}[t]{0.44\textwidth}
\vspace{0pt}
\centering
\begin{minipage}{\linewidth}
\centering
\footnotesize
\setlength{\tabcolsep}{3pt}
\begin{tabular}{>{\raggedright\arraybackslash}p{\dimexpr.68\linewidth-2\tabcolsep\relax}
>{\raggedleft\arraybackslash}p{\dimexpr.32\linewidth-2\tabcolsep\relax}}
\toprule
\appendixcolhead{Method} & \appendixcolhead{Driving Score} \\
\midrule
AD-MLP \citeyearpar{bench2drive} & 18.05 \\
TCP \citeyearpar{tcp} & 40.70 \\
UniAD-Tiny \citeyearpar{uniad} & 40.73 \\
VAD \citeyearpar{vad} & 42.35 \\
UniAD-Base \citeyearpar{uniad} & 45.81 \\
TCP-traj w/o distillation \citeyearpar{tcp} & 49.30 \\
TCP-traj \citeyearpar{tcp} & 59.90 \\
WoTE \citeyearpar{wote} & 61.71 \\
ThinkTwice \citeyearpar{thinktwice} & 62.44 \\
DriveAdapter \citeyearpar{driveadapter} & 64.22 \\
Drive-JEPA \citeyearpar{drivejepa} & 64.52 \\
SafeDrive \citeyearpar{safedrive} & 66.77 \\
ReCogDrive \citeyearpar{recogdrive} & \underline{71.36} \\
\rowcolor{TableOurs}\textbf{Ours} & \textbf{\bTwoDOurs} \\
\bottomrule
\end{tabular}
\captionsetup{font=small,width=\linewidth,justification=RaggedRight,singlelinecheck=false}
\captionof{table}{\textbf{Extended Bench2Drive comparison.} Driving Score.}
\label{tab:b2d-ds}
\end{minipage}

\end{minipage}
\par\smallskip

Route re-ranking uses the 1\,m \texttt{set\_global\_plan} polyline; command
retention uses the
reference agents' 50\,m waypoints~\citep{bench2drive}. Ego pose, velocity, and
acceleration come from CARLA; the reference agents estimate these quantities from GNSS and IMU measurements.

To avoid map-streaming crashes, we load the map within 4000\,m on
routes 11715, 11755, 23659, 23695, 23708, 24041, and 24071, and 2000\,m on
route 3800.
\par\smallskip

\FloatBarrier

\end{document}